\documentclass[11pt]{article}

\usepackage[final]{acl}

\usepackage{times}
\usepackage{latexsym}

\usepackage[T1]{fontenc}

\usepackage[utf8]{inputenc}

\usepackage{microtype}

\usepackage{inconsolata}

\usepackage{graphicx}

\usepackage{CJKutf8}
\usepackage{amssymb}
\usepackage{amsmath} 
\usepackage{multirow}

\usepackage{booktabs}
\usepackage{dashrule}
\usepackage{color}
\usepackage{xcolor}
\usepackage{graphicx}
\usepackage{caption}
\usepackage{threeparttable}
\usepackage[table]{xcolor}
\usepackage{array}
\usepackage{tabularx}
\usepackage{arydshln}
\usepackage[export]{adjustbox}
\usepackage{ragged2e}
\usepackage{floatflt}
\usepackage{makecell}
\usepackage[dvipsnames]{xcolor}

\newcommand{\dashedmidrule}{%
  \noalign{\vskip\aboverulesep}%
  \hdashline
  \noalign{\vskip\belowrulesep}%
}
\newcommand{\cjkbold}[1]{{\fontseries{bx}\selectfont #1}}

\let\oldcolorbox\colorbox

\renewcommand{\colorbox}[2]{%
  \begingroup%
    \setlength{\fboxsep}{0pt}%
    \oldcolorbox{#1}{\strut #2}%
  \endgroup%
}

\title{FEA-SLT: A Gloss-Free End-to-End Framework for Facial-Expression-Aware Sign Language Translation}

\author{
 Guobin Tu\textsuperscript{1} \and Di Weng\textsuperscript{1,$\dagger$}
\\ 
\textsuperscript{1}School of Software Technology, Zhejiang University\\
\textsuperscript{$\dagger$}Corresponding author \\
\texttt{\{tuguobin, dweng\}@zju.edu.cn}
}

\newcounter{ablationrow}
\newcommand{\ablnum}{\stepcounter{ablationrow}(\arabic{ablationrow})}

\begin{document}
\begin{CJK}{UTF8}{gbsn}

\maketitle

\begin{abstract}

Sign Language Translation (SLT) is a challenging cross-modal task requiring joint modeling of manual articulations and non-manual signals.
Existing gloss-free SLT methods effectively capture gestural dynamics but often underutilize facial expressions, which play crucial grammatical and disambiguating roles.
This limitation can cause semantic degradation when distinct concepts share similar manual configurations.
To address this issue, we propose FEA-SLT (\textbf{F}acial-\textbf{E}xpression-\textbf{A}ware \textbf{S}ign \textbf{L}anguage \textbf{T}ranslation), a gloss-free end-to-end framework that uses facial dynamics to provide complementary cues to manual signals.
FEA-SLT employs a domain-transferred facial encoder to extract expression-sensitive representations and integrates them with manual features through a linguistically motivated \textit{Facial-Expression-Aware Fusion} (FEAF) module.
FEAF captures reciprocal dependencies between manual and facial channels via bidirectional modulation, enhancing syntactic fidelity.
Experiments on PHOENIX14T and CSL-Daily show that FEA-SLT achieves state-of-the-art BLEU performance among gloss-free methods, while targeted analyses support improved translation of facial-sensitive utterances.
Code is available at ~\url{https://github.com/TuGuobin/FEA-SLT}.

\end{abstract}

\section{Introduction}
\label{sec:introduction}

Sign language (SL) is the primary communication modality for over 70 million deaf people worldwide~\cite{wfd2026faqs}. 
It is a sophisticated semiotic system encoding linguistic information through two complementary channels: Manual Signals (MS), comprising hand shapes and movements, and Non-Manual Signals (NMS), encompassing facial expressions, mouthing, and head positioning~\cite{pfau2012sign,rastgoo2022all}. 
NMS serve a dual linguistic purpose: they provide grammatical structure (e.g., distinguishing interrogatives from declaratives) and embed semantic cues that shape how MS are interpreted~\cite{elliott2013facial,reilly1992affective}.

\begin{figure}[t]
\centering
\includegraphics[width=\columnwidth]{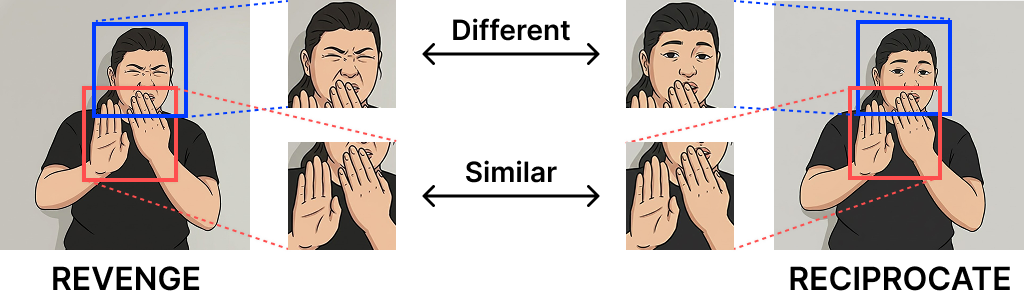}
\caption{Facial cues accompanying visually similar manual signs. Examples from Chinese Sign Language where signs with nearly identical manual articulations (e.g., ``REVENGE'' vs. ``RECIPROCATE'') can require contrasting facial configurations. Portraits are stylized by AI for privacy compliance.}
\label{fig:motivation}
\end{figure}

Recent advances in Sign Language Translation (SLT) have established effective paradigms for bridging the communication gap between Deaf and hearing communities~\cite{lin2023glofe, rust2024towards, gueuwou2025shubert, jiang2026think}.
However, current SLT approaches often focus on holistic visual representations, lacking explicit modeling of facial cues. 
This limitation leads to substantial degradation in semantic fidelity, particularly when processing signs whose meanings depend strongly on facial cues.
For instance, in Chinese Sign Language, ``REVENGE'' and ``RECIPROCATE'' (Figure~\ref{fig:motivation}) share similar hand movements and can require contrasting facial configurations for disambiguation~\cite{nmfs2021hu}.
Neglecting these cues can lead to incorrect translations.

Existing approaches have explored multi-stream architectures and emotion-aware datasets to mitigate manual-centric bias. 
Nevertheless, these methods generally do not combine explicit facial-expression extraction with dedicated cross-modal fusion between facial dynamics and MS.
Available datasets also remain constrained by limited scale and prohibitive annotation complexity.
These limitations can hinder the modeling of expression-dependent linguistic phenomena when distinct concepts share identical MS.
Furthermore, generic vision encoders pretrained on broad tasks often overlook the subtle muscle dynamics that constitute grammatical facial markers in SL.
These limitations motivate the development of dedicated facial-expression encoders capable of modeling fine-grained muscle dynamics beyond generic visual representations, together with cross-modal fusion mechanisms that integrate facial and manual features.

Thus, we propose \textbf{FEA-SLT} (\textbf{F}acial-\textbf{E}xpression-\textbf{A}ware \textbf{S}ign \textbf{L}anguage \textbf{T}ranslation), a gloss-free end-to-end framework that explicitly models facial expressions as complementary cues to manual signals.
FEA-SLT consists of a multi-stream architecture that processes spatial configurations, motion dynamics, and facial-expression features separately before fusing them with bidirectional prosodic modulation. 
Our key insight is that facial expressions, as one of the most readily extractable components of NMS, provide expression-sensitive cues that complement MS.
Their conceptual alignment with Facial Expression Recognition (FER) objectives provides a practical pathway for knowledge transfer. By adapting FER priors, our framework emphasizes expression-sensitive facial dynamics without requiring SL-specific facial annotations.
Our contributions are summarized as follows:

\begin{itemize}
    \item We introduce a decoupled multi-path architecture that processes manual and facial channels through dedicated encoders, leveraging domain-transferred facial representations to capture expression-sensitive dynamics important for grammatical disambiguation. 
    \item We propose a \textit{Facial-Expression-Aware Fusion} (FEAF) module that explicitly models the prosodic coordination between manual and facial channels via bidirectional modulation, helping preserve facial cues that single-stream or unidirectional approaches may underuse.
    \item Extensive evaluations demonstrate that FEA-SLT achieves state-of-the-art performance among gloss-free methods on BLEU metrics while remaining competitive on ROUGE-L, with ablation studies and targeted subset analyses showing benefits from domain-transferred facial features and bidirectional fusion, particularly for utterances involving affective or interrogative semantics.
\end{itemize}

\section{Related Work}
\label{sec:related_work}

\subsection{Gloss-Free Sign Language Translation}
\label{subsec:gloss_free_slt}

SLT has progressively shifted from gloss-based pipelines~\cite{camgoz2018neural,camgoz2020sign,zhou2021improving} toward end-to-end gloss-free frameworks to circumvent costly manual annotations and inherent information bottlenecks~\cite{jin2022prior, chen2022simple, zhang2023sltunet}.
Early direct mapping approaches struggled with the pronounced visual-linguistic modality gap, prompting the adoption of vision-language pretraining and retrieval-based paradigms to enhance cross-modal semantic consistency~\cite{zhao2022conditional,zhou2023glofe,jiao2024vap}. 
Recently, large language models (LLMs)~\cite{openai2024gpt4technicalreport} and multimodal LLMs (MLLMs) have been leveraged to bridge this gap by projecting visual features into text-like latent spaces or generating intermediate textual descriptions~\cite{chen2024fla,wong2024sign2gpt,gong2024llms,asasi2025beyondgloss}. 
Despite substantial gains in lexical accuracy, contemporary gloss‑free methods remain heavily MS‑centric. 
They predominantly treat facial expressions and other NMS as auxiliary visual context rather than integral grammatical components, which can obscure distinctions that depend on facial cues.

\begin{figure*}[!t]
\centering
\includegraphics[width=\textwidth]{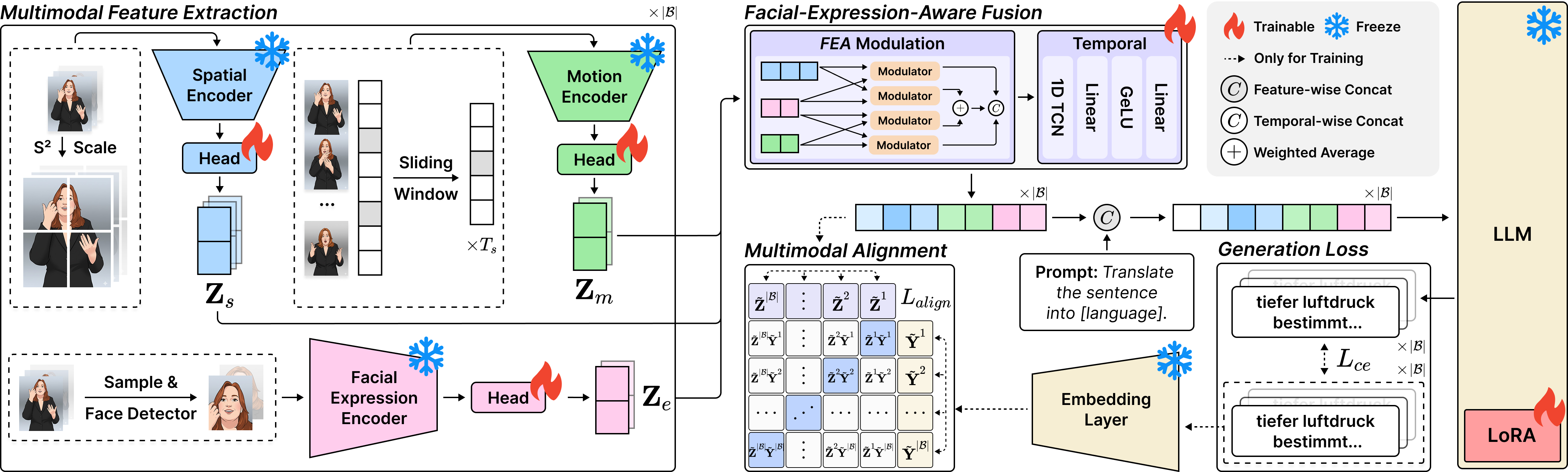}
\caption{Overview of the FEA-SLT framework. The end-to-end pipeline processes sign language videos through three stages: (i) \textbf{Multimodal Feature Extraction} disentangles spatial, motion, and facial-expression features using dedicated encoders; (ii) \textbf{Facial-Expression-Aware Fusion (FEAF)} integrates these representations via a bidirectional conditioner that promotes prosodic co-articulation; (iii) \textbf{Translation Generation} conditions an LLM on fused features to produce target text.}
\label{fig:framework}
\end{figure*}

\subsection{Facial-Expression Modeling}
\label{subsec:nms_facial}

Linguistic research establishes facial expressions as indispensable syntactic markers in SL, governing grammatical structures such as interrogatives and negations while engaging in bidirectional prosodic coordination with MS. 
Specifically, facial dynamics prosodically constrain gestural velocity and amplitude, whereas concurrent MS reciprocally shape the temporal realization of facial markers~\cite{elliott2013facial,brentari2002prosody,pfau2010nonmanuals,viegas2023including,sharma2024facial,chua2025perspectives,chua2025emosign}. 
This tightly coupled system supplies syntactic context that can help distinguish visually similar MS.

Explicit computational modeling of these facial cues in SLT remains challenging. Existing methodologies exhibit distinct limitations in capturing facial semantics. 
First, multi-stream architectures explicitly separate face, hand, and body streams, but do not focus on modeling facial temporal evolution for SLT~\cite{gueuwou2025signmusketeers}.
Second, skeleton-based approaches extract geometric landmarks that primarily encode structural poses rather than the semantic variations required for grammatical disambiguation~\cite{jiao2023cosign,li2025unisign}. 
Third, recent MLLM-based methods verbalize facial dynamics for alignment, but discretizing continuous expressions into static tokens can discard fine-grained intensity and temporal progression~\cite{kim2025mmslt}.
While recent datasets like EmoSign~\cite{chua2025emosign} target affective facial cues, they remain constrained by limited scale and annotation complexity.
Therefore, vision encoders pretrained on general‑purpose tasks often lack domain-specific calibration for subtle facial muscle dynamics suitable for SL.

A further limitation across existing frameworks is the limited explicit interaction between facial dynamics and MS.
The interplay between these modalities is inherently synergistic, with facial expressions prosodically modulating the kinematic patterns of manual signs to jointly shape utterance structure. 
Current translation models often treat these signals independently, limiting their ability to model expression-dependent linguistic phenomena.

\section{Methodology}
\label{sec:method}

We propose \textbf{FEA-SLT}, a gloss-free end-to-end framework that introduces a decoupled facial-driven stream alongside a bidirectional modulation mechanism to preserve prosodic coordination, thereby explicitly incorporating facial cues into grammatical and semantic modeling.

\subsection{Framework Architecture}
\label{subsec:framework}

FEA-SLT adopts the strong spatial-motion paradigm exemplified by SpaMo~\cite{hwang2025spamo} as a controlled backbone to investigate whether facial dynamics provide complementary information beyond MS alone.
Our contribution is a FER-transferred facial stream and a bidirectional facial-manual modulation mechanism that integrates expression-sensitive cues with manual spatial-motion representations.
This formulation allows us to isolate the contribution of facial-expression-aware modeling itself under a strong contemporary gloss-free SLT framework.


As illustrated in Figure~\ref{fig:framework}, FEA-SLT processes an input video sequence $\mathbf{X} = \{x_t\}_{t=0}^{T}$ to generate a target translation $\mathbf{Y} = \{y_u\}_{u=0}^{U}$ through three sequential stages:
(i) \textbf{Multimodal Feature Extraction}: Disentangling the input $\mathbf{X}$ into spatial ($\mathbf{Z}_s$), motion ($\mathbf{Z}_m$), and facial-expression ($\mathbf{Z}_e$) representations;
(ii) \textbf{Facial-Expression-Aware Fusion (FEAF)}: Establishing bidirectional cross-modal modulation, followed by a temporal layer for short-term modeling to yield the final multimodal representation $\mathbf{Z}$;
(iii) \textbf{Translation Generation}: Conditioning an LLM on the fused representations $\mathbf{Z}$ to produce the target sequence $\mathbf{Y}$.

\subsection{Multimodal Feature Extraction}
\label{subsec:feature_extraction}

To capture the dual-channel nature of SL, we employ decoupled feature extractors that align with this fundamental linguistic structure. Specifically, we utilize dedicated extractors for spatial configuration and motion dynamics to represent MS, alongside a facial stream to capture facial expressions. 
This design reflects how MS convey core lexical meaning while facial expressions simultaneously encode grammatical functions~\cite{pfau2012sign,chua2025emosign}.

\paragraph{Spatial Feature Extraction.} 
Spatial features encode static configurations such as hand shapes and body postures. 
To mitigate information loss in small regions, we employ a multi-scale strategy ($S^2$)~\cite{shi2024when} that processes each frame at two resolutions: global context ($224^2$) and local fine-grained details ($448^2$). 
The high-resolution input is partitioned into four patches to fit the encoder. 
A Vision Transformer (ViT)~\cite{dosovitskiy2021image} extracts the [CLS] token from each view (1 global, 4 local). 
We then aggregate these representations into $\hat{\mathbf{z}}_{s}^t$ for each frame $x_t$ by concatenating the global view's [CLS] token with the average of the four local [CLS] tokens.

\paragraph{Motion Feature Extraction.} 
Motion features capture the temporal dynamics and kinematic variations of signs. 
We segment videos into overlapping clips via a sliding window of width $w$ and stride $s_m$. 
A pretrained video encoder extracts features $\hat{\mathbf{Z}}_m$ from each clip, capturing motion dynamics.

\paragraph{Facial-Expression Feature (FE) Extraction.} 
To encode expression-sensitive facial dynamics, we leverage a ViT pretrained on the FER dataset as a dedicated feature extractor.
Although trained for categorical facial expression classification, the FER encoder learns expression-sensitive latent representations that preserve fine-grained temporal variations in facial muscle dynamics. 
These representations share kinematic primitives with the prosodic markers of SL (detailed in Appendix~\ref{sec:domain}), enabling extraction of continuous expression dynamics without SL-specific annotation.
This domain transfer provides an expression-sensitive prior without requiring sign-specific facial labels.

To mitigate temporal redundancy inherent in slow-evolving facial dynamics, we uniformly downsample frames at interval $s_e$ and extract aligned facial Regions of Interest (ROIs). 
These ROIs are subsequently encoded by the FER encoder, producing a sequence of frame-level facial features. 
Feature-level interpolation bridges occasional detection gaps while respecting the prosodic pace of facial markers (Appendix~\ref{sec:failure}), yielding a smooth sequence $\hat{\mathbf{Z}}_e$ that faithfully preserves the FER encoder's expression-sensitive latent manifold.

Subsequently, all features are mapped into a unified dimension $d$ through a lightweight head layer:
\begin{equation}
\mathbf{Z}_s \in \mathbb{R}^{T \times d}, \mathbf{Z}_m \in \mathbb{R}^{T_m \times d}, \mathbf{Z}_e \in \mathbb{R}^{T_e \times d},
\end{equation}
where $T_m$ and $T_e$ denote the sequence lengths of motion and facial-expression features, respectively.

\begin{figure}[!t]
\centering
\includegraphics[width=\columnwidth]{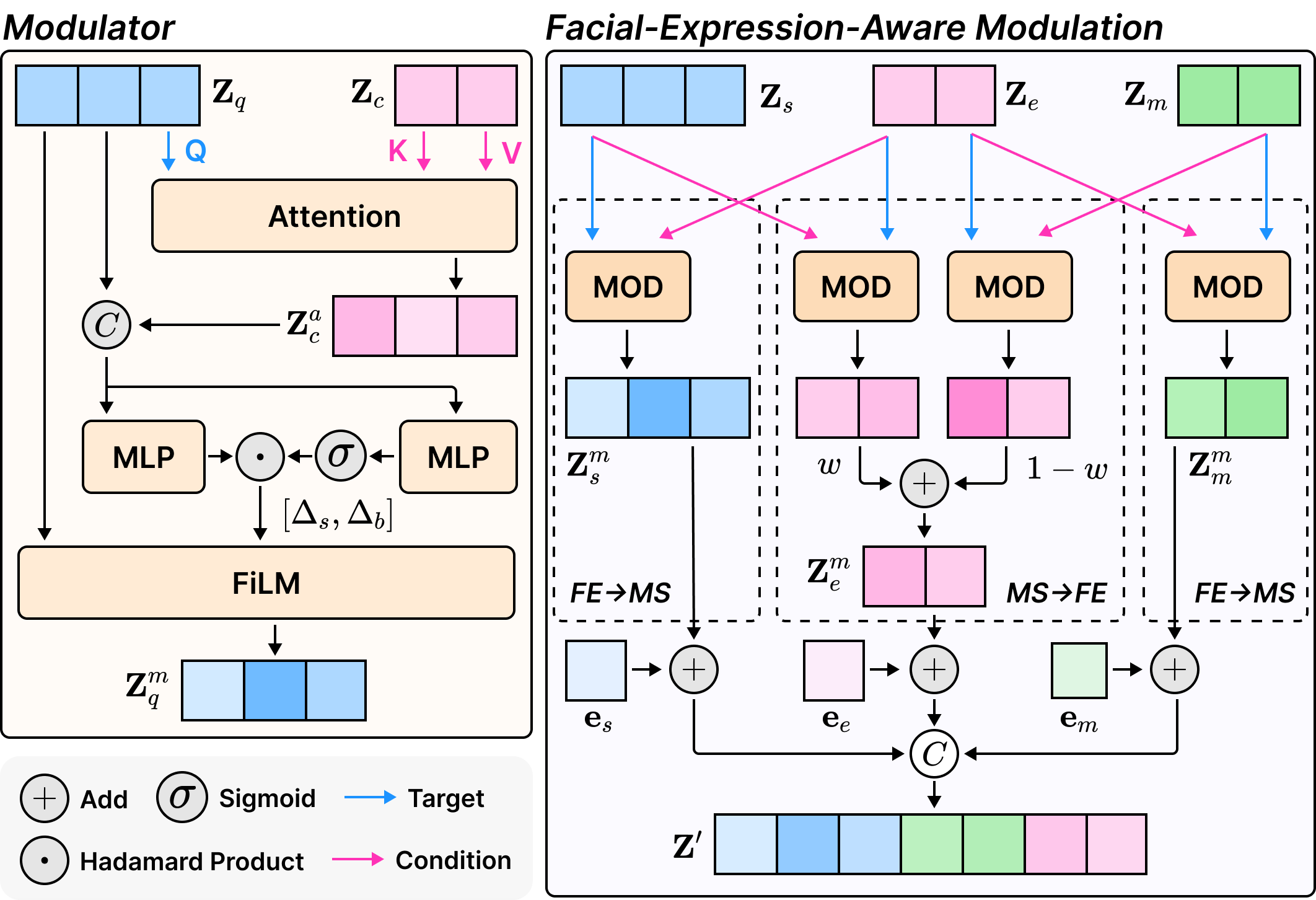}
\caption{Architecture of the \textbf{Modulator} and \textbf{Facial-Expression-Aware Modulation (FEAM)} module. The Modulator dynamically predicts scaling and gating parameters to modulate cross-modal features, while FEAM leverages bidirectional modulation to enable mutual refinement between facial and manual streams. In FEAM, blue arrows indicate target streams and pink arrows indicate conditioning streams. The two FE\(\rightarrow\)MS branches modulate the spatial and motion streams, respectively, while the central MS\(\rightarrow\)FE branch aggregates spatial- and motion-conditioned facial features before concatenation.}
\label{fig:feam}
\end{figure}

\subsection{Facial-Expression-Aware Fusion (FEAF)}
\label{subsec:feaf}

While we decouple MS and FE at the extraction stage to employ domain-specific encoders, we re-integrate them through bidirectional modulation at the fusion stage to model their linguistic interdependence. 
The FEAF module consists of the FEAM and the Temporal Layer, among which the key component of FEAM is the Modulator (MOD).

\paragraph{Modulator.} 
As shown in Figure~\ref{fig:feam}, the Modulator establishes frame-wise temporal correspondence and predicts adaptive transformation parameters to modulate target features.

Given target features $\mathbf{Z}_q \in \mathbb{R}^{T_q \times d}$ and conditioning features $\mathbf{Z}_c \in \mathbb{R}^{T_c \times d}$, the modulation $\mathbf{Z}_q^{\text{m}} = \operatorname{MOD}(\mathbf{Z}_q, \mathbf{Z}_c)$ first aligns $\mathbf{Z}_c$ to the temporal resolution of $\mathbf{Z}_q$ via cross-attention:
\begin{equation}
\mathbf {A} = \operatorname {softmax}\left ( \frac {\mathbf {Z}_q W_q W_k^\top \mathbf {Z}_c^\top}{\sqrt {d}} \right), \mathbf {Z}_c^{\text {a}} = \mathbf {A} \mathbf {Z}_c W_v,
\end{equation}
where $W_q, W_k, W_v \in \mathbb{R}^{d \times d}$ are learnable projections and $\mathbf{Z}_c^{\text{a}} \in \mathbb{R}^{T_q \times d}$ denotes the aligned conditioning features.
Subsequently, the joint context $[\mathbf{Z}_q, \mathbf{Z}_c^{\text{a}}]$ is processed by two parallel projection heads. A parameter head predicts channel-wise affine shifts, while a gating head estimates frame-level modulation intensity:
\begin{align}
[\tilde{\Delta}_s, \tilde{\Delta}_b] &= \operatorname{MLP}_{\text{param}}\big( [\mathbf{Z}_q, \mathbf{Z}_c^{\text{a}}] \big), \\
\mathbf{g} &= \sigma\left( \operatorname{MLP}_{\text{gate}}\big( [\mathbf{Z}_q, \mathbf{Z}_c^{\text{a}}] \big) \right), \\
[\Delta_s, \Delta_b] &= \alpha \cdot \tanh([\tilde{\Delta}_s, \tilde{\Delta}_b]) \odot \mathbf{g},
\end{align}
where $[\cdot, \cdot]$ denotes feature-wise concatenation, $\Delta_s, \Delta_b \in \mathbb{R}^{T_q \times d}$, and $\alpha$ controls the modulation magnitude. 
The modulated output is obtained by FiLM-style transformation~\cite{perez2018film}:
\begin{equation}
\mathbf{Z}_q^{\text{m}} = \mathbf{Z}_q \odot (\mathbf{1} + \Delta_s) + \Delta_b.
\end{equation}

\paragraph{Facial-Expression-Aware Modulation (FEAM).} 
To model the reciprocal dependencies among spatial ($\mathbf{Z}_s$), motion ($\mathbf{Z}_m$), and facial-expression ($\mathbf{Z}_e$) representations, FEAM employs a bidirectional prosodic modulation strategy that adaptively fuses cross-modal contexts into a unified sequential input. Forward modulation conditions the spatial and motion streams on facial context:
\begin{equation}
\begin{aligned}
  \mathbf{Z}_s^{\text{m}} &= \operatorname{MOD}(\mathbf{Z}_s, \mathbf{Z}_e), \\
  \mathbf{Z}_m^{\text{m}} &= \operatorname{MOD}(\mathbf{Z}_m, \mathbf{Z}_e),
\end{aligned}
\end{equation}
while reverse modulation refines facial features conditioned on spatial and motion dynamics:
\begin{equation}
\begin{aligned}
  \mathbf{Z}_e^{(s)} &= \operatorname{MOD}(\mathbf{Z}_e, \mathbf{Z}_s), \\
  \mathbf{Z}_e^{(m)} &= \operatorname{MOD}(\mathbf{Z}_e, \mathbf{Z}_m).
\end{aligned}
\end{equation}

The reverse-modulated facial features are aggregated via a learnable scalar weight $w_{\text{p}}$ to balance complementary cues:
\begin{equation}
w = \sigma(w_{\text{p}}), \mathbf{Z}_e^{\text{m}} = w \cdot \mathbf{Z}_e^{(s)} + (1 - w) \cdot \mathbf{Z}_e^{(m)}.
\end{equation}

Each modality-specific representation is subsequently augmented with a learnable class embedding $\mathbf{e}_k$ ($k \in \{s, m, e\}$) to introduce global semantic constraints: $\tilde{\mathbf{Z}}_k = \mathbf{Z}_k^{\text{m}} + \mathbf{e}_k$. The enhanced features are concatenated along the temporal axis to yield the unified sequence $\mathbf{Z}' = [\tilde{\mathbf{Z}}_s; \tilde{\mathbf{Z}}_m; \tilde{\mathbf{Z}}_e]$.

\paragraph{Temporal Layer.} For short-term temporal modeling, the fused sequence $\mathbf{Z}'$ is processed by a 1D Temporal Convolutional Network (TCN)~\cite{bai2018tcn}, followed by a GELU-activated MLP that projects the features into the embedding space of the target language model:
\begin{equation}
\mathbf{Z} = \operatorname{MLP}\left(\operatorname{TCN}\left( \mathbf{Z}' \right) \right) \in \mathbb{R}^{L \times d_{\text{llm}}},
\end{equation}
where $L$ denotes the downsampled sequence length and $d_{\text{llm}}$ corresponds to the hidden dimension of the language model. Implementation details of the TCN architecture are provided in Appendix \ref{sec:temporal_modeling}.

By explicitly conditioning manual representations on facial dynamics and vice versa, FEAF preserves grammatical and facial-sensitive cues prior to autoregressive generation.

\subsection{Training Details}
\label{subsec:training}

\paragraph{Multimodal Alignment (MA).} To bridge the semantic gap between multimodal visual features and target text, we apply a bidirectional contrastive loss $\mathcal{L}_{\text{align}}$ between global multimodal representations $\tilde{\mathbf{Z}}=\operatorname{MeanPool}(\mathbf{Z})$ and global target text representations $\tilde{\mathbf{Y}}=\operatorname{MeanPool}(\operatorname{Embedding}(\mathbf{Y}))$ over mini-batch $\mathcal{B}$ \cite{radford2021learning, zhou2023glofe}:
\begin{equation}
\begin{aligned}
\mathcal{L}_{\text{align}}
=-\frac{1}{2|\mathcal{B}|}\sum_{i=1}^{|\mathcal{B}|}\Bigg[
&\underbrace{\log\frac{\exp(s_{ii}/\tau)}
{\sum_{j=1}^{|\mathcal{B}|}\exp(s_{ij}/\tau)}}_{\text{Sign} \to \text{Text}}\\
{}+{}&
\underbrace{\log\frac{\exp(s_{ii}/\tau)}
{\sum_{j=1}^{|\mathcal{B}|}\exp(s_{ji}/\tau)}}_{\text{Text} \to \text{Sign}}
\Bigg],
\end{aligned}
\end{equation}
where $s_{ij}=\operatorname{sim}(\tilde{\mathbf{Z}}^i,\tilde{\mathbf{Y}}^j)$ denotes cosine similarity, $\tau$ is a learnable temperature parameter, and $|\mathcal{B}|$ is the batch size.

\paragraph{Generation Loss.} We utilize an LLM for translation generation. Given a task-specific prompt $\mathbf{P}$ that employs in-context learning \cite{brown2020language} (detailed in Appendix \ref{sec:prompt_design}), alongside the fused features $\mathbf{Z}$ as conditional inputs, the LLM predicts the target sequence $\mathbf{Y}$ via teacher forcing. This is optimized using cross-entropy loss with label smoothing \cite{szegedy2016rethinking}:
\begin{equation}
\begin{aligned}
\mathcal{L}_{\text{ce}} = -\frac{1}{|\mathcal{B}|}
\sum_{i=1}^{|\mathcal{B}|}\sum_{u=1}^{U}
\log P\big(y_u^i \mid y_{<u}^i, [\mathbf{Z}^i,\mathbf{P}]\big).
\end{aligned}
\end{equation}
where $P(\cdot)$ represents the LLM's token probability distribution, and $y_{<u}^i$ denotes the preceding token sequence for sample $i$.

Unlike prevalent multi-stage pipelines constrained by heavy pre-training overhead, FEA-SLT adopts a single-stage paradigm, jointly optimizing both objectives in an end-to-end manner:
\begin{equation}
\mathcal{L} = \mathcal{L}_{\text{ce}} + \lambda \mathcal{L}_{\text{align}},
\end{equation}
where $\lambda$ balances the alignment and generation objectives. To ensure computational efficiency while preserving generative capacity, we employ Low-Rank Adaptation (LoRA) \cite{hu2022lora} for parameter-efficient fine-tuning.

\begin{table*}[ht]
  \centering
  \renewcommand{\arraystretch}{0.7}
  \adjustbox{width=\textwidth}{
    \begin{tabular}{lccccccccccc}
  \toprule
      \multirow{2}{*}{\textbf{Method}} & \multicolumn{5}{c}{\textbf{PHOENIX14T}} & \multicolumn{5}{c}{\textbf{CSL-Daily}} \\
      & B-1 & B-2 & B-3 & B-4 & R-L & B-1 & B-2 & B-3 & B-4 & R-L \\
      \midrule
      
      \multicolumn{11}{c}{\cellcolor{gray!10}\textbf{Gloss-based}} \\
      \midrule
      SLRT \cite{camgoz2020sign} & 46.61 & 33.73 & 26.19 & 21.32 & -- & 37.38\textsuperscript{*} & 24.36\textsuperscript{*} & 16.55\textsuperscript{*} & 11.79\textsuperscript{*} & 36.74\textsuperscript{*} \\
      BN-TIN-Transf+SignBT \cite{zhou2021improving} & 50.80 & 37.75 & 29.72 & 24.32 & 49.54 & 51.42 & 37.26 & 27.76 & 21.34 & 49.31 \\
      MMTLB \cite{chen2022simple} & 53.97 & 41.75 & 33.84 & 28.39 & 52.65 & 53.31 & 40.41 & 30.87 & 23.92 & 53.25 \\
      TS-SLT \cite{chen2022ts-slt} & 54.90 & 42.43 & 34.46 & 28.95 & 53.48 & 55.44 & 42.59 & 32.87 & 25.79 & 55.72 \\
      SLTUNET \cite{zhang2023sltunet} & 52.92 & 41.76 & 33.99 & 28.47 & 52.11 & 54.98 & 41.44 & 31.84 & 25.01 & 54.08 \\
      \midrule
      
      \multicolumn{11}{c}{\cellcolor{gray!10}\textbf{Weakly supervised gloss-free}} \\
      \midrule
      TSPNet \cite{li2020tspnet} & 36.10 & 23.12 & 16.88 & 13.41 & 34.96 & 17.09\textsuperscript{\textdagger} & 8.98\textsuperscript{\textdagger} & 5.07\textsuperscript{\textdagger} & 2.97\textsuperscript{\textdagger} & 18.38\textsuperscript{\textdagger} \\
      GASLT \cite{yin2023gloss} & 39.07 & 26.74 & 21.86 & 15.74 & 39.86 & 19.90 & 9.94 & 5.98 & 4.07 & 20.35 \\
      ConSLT \cite{fu2023conslt} & -- & -- & -- & 21.59 & 47.69 & -- & -- & -- & 14.53 & 40.98 \\
      VAP \cite{jiao2024vap} & 53.07 & -- & -- & 26.16 & 51.28 & 49.99 & -- & -- & 20.85 & 48.56 \\
      \midrule
      
      \multicolumn{11}{c}{\cellcolor{gray!10}\textbf{Gloss-free}} \\
      \midrule
      NSLT +Luong \cite{luong2015effective} & 29.86 & 17.52 & 11.96 & 9.00 & 30.70 & 34.16 & 19.57 & 11.84 & 7.56 & 34.54 \\
      GFSLT-VLP \cite{zhou2023glofe} & 43.71 & 33.18 & 26.11 & 21.44 & 42.49 & 39.37 & 24.93 & 16.26 & 11.00 & 36.44 \\
      FLa-LLM \cite{chen2024fla} & 46.29 & 35.33 & 28.03 & 23.09 & 45.27 & 37.13 & 25.12 & 18.38 & 14.20 & 37.25 \\
      Sign2GPT \cite{wong2024sign2gpt} & 49.54 & 35.96 & 28.83 & 22.52 & \underline{48.90} & 41.75 & 28.73 & 20.60 & 15.40 & 42.36 \\
      SignLLM \cite{gong2024llms} & 45.21 & 34.78 & 28.05 & 23.40 & 44.49 & 39.55 & 28.13 & 20.07 & 15.75 & 39.91 \\
      MLSLT \cite{tan2025mlslt} & -- & -- & -- & 24.23 & \textbf{50.60} & -- & -- & -- & 14.18 & 40.00 \\
      MMSLT \cite{kim2025mmslt} & 48.92 & \underline{38.12} & \underline{30.79} & \underline{25.73} & 47.97 & \underline{49.87} & 36.37 & \underline{27.29} & \underline{21.11} & \underline{48.92} \\

      SpaMo \cite{hwang2025spamo} & \underline{49.80} & 37.32 & 29.50 & 24.32 & 46.57 & 48.90 & \underline{36.90} & 26.78 & 20.55 & 47.46 \\
      \dashedmidrule

      \textbf{FEA-SLT (Ours)} & \textbf{52.59} & \textbf{39.72} & \textbf{31.67} & \textbf{26.38} & 48.63 & \textbf{50.83} & \textbf{37.82} & \textbf{29.03} & \textbf{22.94} & \textbf{50.40} \\
    \bottomrule
    \end{tabular}
  }
  
  \caption{BLEU-1 to BLEU-4 and ROUGE-L results on PHOENIX14T and CSL-Daily datasets (Test Set). 
  The best results for gloss-free models are in \textbf{bold}, while the second-best are \underline{underlined}. Missing values are denoted by --. \textsuperscript{*} and \textsuperscript{\textdagger} denote reproduced results~\cite{zhou2023glofe,yin2023gloss}.}
  \label{tab:sota}
\end{table*}

\section{Experiments}
\label{sec:experiments}

\subsection{Datasets}
\label{subsec:datasets}

We evaluate our proposed FEA-SLT on two widely adopted SLT benchmarks: 
\textbf{PHOENIX14T} \cite{camgoz2018neural}, a German Sign Language (DGS) dataset focused on weather forecasts, which consists of 7,096 training, 519 validation, and 642 test samples characterized by rich NMS;
\textbf{CSL-Daily} \cite{zhou2021improving}, a large-scale Chinese Sign Language (CSL) dataset covering diverse daily scenarios, comprising 18,401 training, 1,077 validation, and 1,176 test samples.
Further details are provided in Appendix \ref{sec:more_dataset_details} and \ref{sec:preprocess_details}.

\subsection{Evaluation Metrics}
\label{subsec:metrics}

To comprehensively assess translation quality, we employ standard metrics: BLEU-1 to BLEU-4 (B-1 to B-4) \cite{papineni2002bleu} for n-gram overlap, ROUGE-L (R-L) \cite{lin2004rouge} for fluency, and BLEURT \cite{sellam2020bleurt} to measure semantic adequacy. 
For more implementation details of evaluation metrics, please refer to Appendix \ref{sec:metrics_details}.
      
\subsection{Experimental Setup}
\label{subsec:setup}

Our FEA-SLT framework is implemented with PyTorch~\cite{pytorch2019paszke}. 
For the spatial and motion encoders, we adopt the pretrained CLIP-ViT-L/14~\cite{radford2021learning} and VideoMAE-L/16~\cite{tong2022videomae} as backbones respectively, following the prior work ~\cite{hwang2025spamo}. 
The facial-expression extraction module employs RetinaFace~\cite{deng2019retinaface} for face detection, and subsequently leverages a ViT fine-tuned on the FER2013 dataset~\cite{goodfellow2013challenges} for facial dynamics representation extraction. 
During the training process, all pre-trained encoder weights remain frozen. 
Only the lightweight head layers, the FEAF module, and the LoRA adapters for the LLM are trainable. 
For the language backbone, we utilize Flan-T5-XL~\cite{chung2022flant5} for PHOENIX14T and mT5-XL~\cite{xue2021mt5} for CSL-Daily, respectively. 
Pretrained weights and additional implementation specifics are provided in Appendix~\ref{sec:preprocess_details} and~\ref{sec:hyperparameters}.

\subsection{Comparison with State-of-the-Art}
\label{subsec:sota}

We conduct a comprehensive evaluation of FEA-SLT against representative gloss-based, weakly supervised, and gloss-free SLT approaches under consistent experimental settings, as summarized in Table~\ref{tab:sota}. To ensure a fair comparison, we exclude methods that leverage external corpora (e.g., Uni-Sign~\cite{li2025unisign}).

\paragraph{Performance on PHOENIX14T.} 
FEA-SLT achieves state-of-the-art performance among gloss-free methods across all BLEU metrics on PHOENIX14T. 
Specifically, it surpasses SpaMo~\cite{hwang2025spamo} by +2.79 in BLEU-1 and outperforms MMSLT~\cite{kim2025mmslt} by +1.60, +0.88, and +0.65 in BLEU-2, BLEU-3, and BLEU-4, respectively. 
These consistent gains across n‑gram orders indicate improvements in both unigram-level and longer n‑gram overlap.
Regarding ROUGE-L, FEA-SLT attains a competitive score of 48.63, trailing MLSLT~\cite{tan2025mlslt} and Sign2GPT~\cite{wong2024sign2gpt} while improving over MMSLT and SpaMo, indicating robust performance in preserving semantic fidelity.

\paragraph{Performance on CSL-Daily.}
On the CSL-Daily benchmark, FEA-SLT establishes new state-of-the-art results among gloss-free methods across all reported metrics. 
Our model outperforms MMSLT by +1.83 in BLEU-4 and +1.48 in ROUGE-L, and also exceeds the weakly supervised VAP~\cite{jiao2024vap} by +2.09 in BLEU-4. 
These results show that the improvements extend to CSL-Daily, a larger benchmark covering diverse daily-life expressions.
We also observe larger improvements on utterances involving affective or interrogative semantics, where facial expressions often play an important communicative role in SL. We provide a diagnostic comparison on a fixed 84-instance facial-sensitive subset in Appendix~\ref{subsec:ambiguity_subset} to further investigate this phenomenon.

\paragraph{Semantic Quality Assessment.}
Standard n-gram metrics may not fully reflect semantic fidelity in cross-modal translation. 
To complement the BLEU and ROUGE analysis, we report BLEURT scores in Table~\ref{tab:bleurt}. 
FEA-SLT attains the best BLEURT scores on both benchmarks, surpassing the strongest prior baseline on each---our reproduced SpaMo and SONAR-SLT~\cite{hamidullah2025sonar}---by +1.9 and +1.4 points, respectively.
These results indicate improved semantic adequacy under BLEURT. When considered alongside the controlled ablations below, they further suggest that facial cues provide complementary information that improves translation quality.

\begin{table}[t]
  \centering
  \renewcommand{\arraystretch}{0.8}
  \resizebox{\linewidth}{!}{
    \begin{tabular}{lcc}
      \toprule
      \textbf{Method} & \textbf{PHOENIX14T} & \textbf{CSL-Daily} \\
      \midrule
      SEM-SLT \cite{hamidullah2024sem} & 48.1 & -- \\
      LiTFiC \cite{jang2025lost} & 52.8 & -- \\
      SONAR-SLT \cite{hamidullah2025sonar} & 54.5 & \underline{56.1} \\
      SpaMo\textsuperscript{*} \cite{hwang2025spamo} & \underline{58.9}\textsuperscript{*} & 53.1\textsuperscript{*} \\
      \dashedmidrule
      \textbf{FEA-SLT (Ours)} & \textbf{60.8} & {\textbf{57.5}} \\
      \bottomrule
    \end{tabular}
  }
  \caption{BLEURT evaluation on PHOENIX14T and CSL-Daily. FEA-SLT outperforms prior work, confirming facial-expression modeling improves semantic fidelity. * denotes reproduced results.}
  \label{tab:bleurt}
\end{table}

\subsection{Ablation Studies}
\label{subsec:ablation}
To rigorously assess the effectiveness of individual components within our framework, we perform extensive ablation studies on the PHOENIX14T test set. 
Additional experimental results and analyses are provided in Appendix \ref{sec:extra-experiments}.

\paragraph{Component Analysis.} 
Table~\ref{tab:ablation} quantifies the contribution of the facial-expression pathway and the directional information flow within FEAM.
We keep the inherited MA objective fixed in the main ablation and treat FEA-SLT without FE and FEAM as the controlled baseline.
Here, MS denotes spatial and motion features: Rows (1)--(5) correspond to the MS-only SpaMo-style baseline, direct FE concatenation, FE\(\rightarrow\)MS, MS\(\rightarrow\)FE, and full bidirectional FEAM, respectively.
In FE\(\rightarrow\)MS, FE conditions the spatial and motion streams before they are concatenated with the original FE; MS\(\rightarrow\)FE reverses this direction and concatenates the modulated FE with the original spatial and motion streams.
The full variant enables both directions, and each resulting stream is augmented with its corresponding class embedding before concatenation.
The controlled baseline achieves 24.76 B-4 and 46.75 R-L.
Incorporating the FE improves B-4 by 0.54 and R-L by 0.58, suggesting that domain-transferred facial features provide additional expression-sensitive information beyond manual-only representations.
Adding bidirectional FEAM further improves B-4 by 1.08 and R-L by 1.30 [(2)\textrightarrow(5)], while the complete facial pathway contributes a statistically significant +1.62 B-4 and +1.88 R-L over the controlled backbone [(1)\textrightarrow(5)] (Appendix~\ref{sec:significance}, $p<0.01$).
These results indicate that the facial pathway supplies expression-sensitive cues beyond manual spatial-motion representations.
Restricting FEAM to a single direction yields suboptimal performance.
Bidirectional mutual modulation outperforms the stronger single-direction variant by 0.63 BLEU-4 [(4)\textrightarrow(5)] and 0.72 ROUGE-L [(3)\textrightarrow(5)], supporting the bidirectional design.

\begin{table}[h]
\centering
\renewcommand{\arraystretch}{0.8}
\setcounter{ablationrow}{0} 
\resizebox{\linewidth}{!}{
    \begin{tabular}{ccccccccc}
    \toprule
     & \textbf{FE} & \textbf{FE$\rightarrow$MS} & \textbf{MS$\rightarrow$FE} & B-1 & B-2 & B-3 & B-4 & R-L \\
    \midrule
    \ablnum & -- & -- & -- & 50.18 & 37.53 & 29.83 & 24.76 & 46.75 \\
    \ablnum & \checkmark & -- & -- & 50.86 & 38.36 & 30.56 & 25.30 & 47.33 \\
    \ablnum & \checkmark & \checkmark & -- & 51.96 & 39.18 & 31.10 & 25.70 & 47.91 \\
    \ablnum & \checkmark & -- & \checkmark & 51.17 & 38.58 & 30.84 & 25.75 & 47.47 \\
    \dashedmidrule
    \ablnum & \checkmark & \checkmark & \checkmark & \textbf{52.59} & \textbf{39.72} & \textbf{31.67} & \textbf{26.38} & \textbf{48.63} \\
    \bottomrule
    \end{tabular}
}
\caption{Ablation study of the facial-expression pathway and directional information flow in FEAM on PHOENIX14T, with MA fixed across rows. MS denotes spatial and motion features.}
\label{tab:ablation}
\end{table}

\paragraph{Interaction with Multimodal Alignment.}
Table~\ref{tab:ma_ablation} separately evaluates the inherited global visual-text MA objective and the facial pathway.
MA alone improves B-4 by 1.68 [(1)\textrightarrow(4)].
Without MA, direct FE concatenation improves B-4 by 1.07 [(1)\textrightarrow(2)] and full FEAM adds 0.60 [(2)\textrightarrow(3)], yielding a total facial-pathway gain of 1.67 [(1)\textrightarrow(3)].
With MA, the corresponding gains are 0.54 [(4)\textrightarrow(5)] and 1.08 [(5)\textrightarrow(6)], totaling 1.62 [(4)\textrightarrow(6)].
These comparable gains indicate that the facial pathway provides complementary benefits beyond global visual-text alignment.

\begin{table}[h]
\centering
\footnotesize
\renewcommand{\arraystretch}{0.8}
\setcounter{ablationrow}{0}
\resizebox{\linewidth}{!}{
    \begin{tabular}{ccccccccc}
    \toprule
     & \textbf{FE} & \textbf{FEAM} & \textbf{MA} & B-1 & B-2 & B-3 & B-4 & R-L \\
    \midrule
    \ablnum & -- & -- & -- & 48.84 & 36.04 & 28.15 & 23.08 & 45.08 \\
    \ablnum & \checkmark & -- & -- & 49.13 & 36.84 & 29.27 & 24.15 & 46.03 \\
    \ablnum & \checkmark & \checkmark & -- & 51.28 & 38.34 & 30.24 & 24.75 & 46.23 \\
    \ablnum & -- & -- & \checkmark & 50.18 & 37.53 & 29.83 & 24.76 & 46.75 \\
    \ablnum & \checkmark & -- & \checkmark & 50.86 & 38.36 & 30.56 & 25.30 & 47.33 \\
    \dashedmidrule
    \ablnum & \checkmark & \checkmark & \checkmark & \textbf{52.59} & \textbf{39.72} & \textbf{31.67} & \textbf{26.38} & \textbf{48.63} \\
    \bottomrule
    \end{tabular}
}
\caption{Ablation study of FE, FEAM, and multimodal alignment (MA) on the PHOENIX14T test set. \checkmark~indicates that a component is activated.}
\label{tab:ma_ablation}
\end{table}

\paragraph{Sensitivity to facial-expression encoders.} 
Table~\ref{tab:facial} analyzes the impact of facial encoder selection while holding the facial-expression pathway and FEAF fusion constant.
This encoder-sensitivity study complements the evidence for facial-expression modeling in Table~\ref{tab:ablation} and Section~\ref{subsec:evaluation}.
The FER-adapted ViT-B/16 yields the strongest results, surpassing its unadapted counterpart by +0.80 BLEU-4 and +0.55 ROUGE-L. 
While the absolute improvements are modest, they remain consistent across evaluation metrics; DINOv2-ViT-B/16 has comparable parameters but performs worse.
These findings suggest that the effectiveness of the facial encoder stems less from generic visual representation power and more from its ability to transfer expression-sensitive features.

\begin{table}[t]
\centering
\setcellgapes{-2pt}
\renewcommand{\arraystretch}{0.8}
\resizebox{\columnwidth}{!}{%
    \begin{tabular}{lcccccc}
    \toprule
    \textbf{Feature Extractor} & \textbf{Params} & \textbf{B-1} & \textbf{B-2} & \textbf{B-3} & \textbf{B-4} & \textbf{R-L} \\
    \midrule
    ResNet-50 & 26M & 51.78 & 39.17 & 31.14 & 25.68 & 48.40 \\
    ViT-B/16 & 86M & 51.44 & 38.88 & 30.93 & 25.58 & 48.08 \\
    DINOv2-ViT-B/16 & 86M & 51.67 & 38.80 & 30.67 & 25.22 & 47.83 \\
    \dashedmidrule
    \textbf{ViT-B/16 (fine-tuned)} & 86M & \textbf{52.59} & \textbf{39.72} & \textbf{31.67} & \textbf{26.38} & \textbf{48.63} \\
    \bottomrule
    \end{tabular}%
}
\caption{Ablation study on facial-expression feature extractors. The FER-adapted encoder outperforms the generic encoders evaluated under the same facial pathway.}
\label{tab:facial}
\end{table}

\subsection{Evaluation}
\label{subsec:evaluation}
To comprehensively assess the efficacy of facial-expression modeling, we conduct quantitative and qualitative analyses on the CSL-Daily benchmark.

\paragraph{Quantitative Analysis.} 
We focus on punctuation prediction for interrogative constructions to evaluate a facial-sensitive form of syntactic disambiguation.
In these cases, facial NMS can contribute to distinctions that are not evident from MS alone.
Following the punctuation-sequence evaluation protocol (Appendix~\ref{subsec:quantitative}), we compute Precision, Recall, and F1 scores on the interrogative subset. 
FEA-SLT outperforms the baseline across all reported metrics, yielding an absolute F1 improvement of 3.69 (Table~\ref{tab:quantitative}). 
These results suggest that explicit modeling of facial dynamics provides useful cues for syntactic distinctions that are difficult to recover from MS alone.

\paragraph{Qualitative Analysis.}
To examine how FEA-SLT handles facial-sensitive translation, we present representative cases from the CSL-Daily test set in Table~\ref{tab:csl_daily_qualitative}.
In the first case, the signer expresses fear, accompanied by a facial configuration with lowered and furrowed eyebrows and tightened eyelids.
The baseline models do not recover this affective content and instead produce semantically unrelated or neutral translations.
In contrast, FEA-SLT recovers the corresponding lexical content and produces a translation closer to the reference.
In the second case, the question word \textit{what} co-occurs with raised eyebrows and an open mouth, which serve as NMS for the interrogative mood. 
SpaMo introduces the unrelated phrase \textit{famous post office}, while FEA-SLT without FE omits the interrogative structure.
FEA-SLT recovers this structure but adds the erroneous ``Premier's,'' showing improved interrogative structure alongside a remaining lexical error.
Additional results for PHOENIX14T and CSL-Daily are provided in Appendix~\ref{subsec:additional_results}.

\definecolor{diffword}{RGB}{255,255,153}
\definecolor{wrong}{RGB}{255,182,193}
\definecolor{correct}{RGB}{152, 251, 152}

\begin{table}[t!]
\centering
\footnotesize
\setlength{\tabcolsep}{0pt}
\renewcommand{\arraystretch}{1.0}
\begin{tabularx}{\columnwidth}{@{}>{\RaggedRight\arraybackslash}X@{}}
\toprule
\hbox{%
  \includegraphics[height=3\baselineskip, keepaspectratio, valign=t]{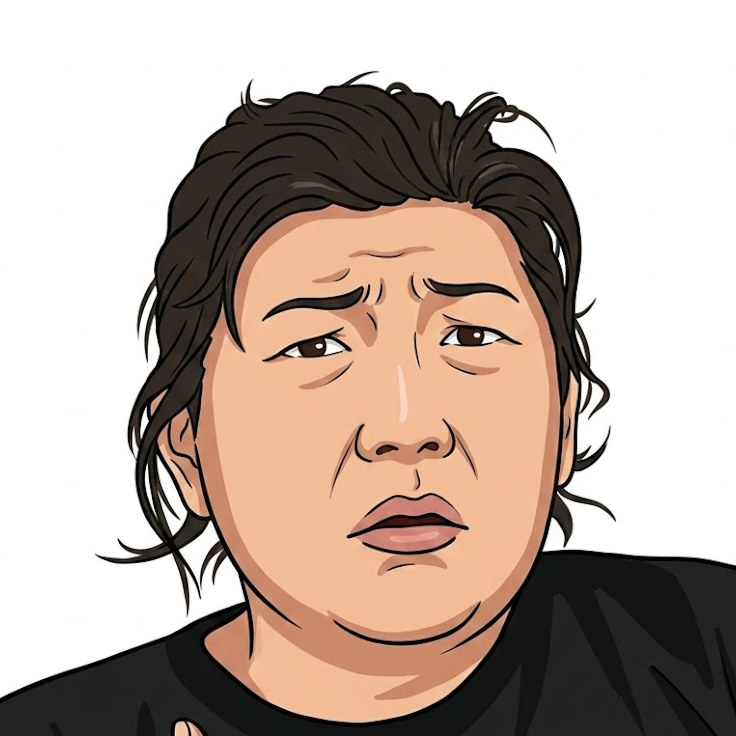}%
  \hspace{4pt}%
  \parbox[t]{\dimexpr\linewidth-0.18\linewidth-8pt\relax}{%
    \textbf{Reference:} \\
    天黑了,我害怕。\\
    (It is dark, I am afraid.)
  }
}
\textbf{SpaMo:} 
\colorbox{wrong}{白天}\colorbox{wrong}{照亮}我。
(\colorbox{wrong}{Daytime illuminates} \colorbox{diffword}{me}.)\\[0.2em]
\textbf{FEA-SLT (w/o FE):} 
\mbox{\colorbox{diffword}{晚上}\colorbox{wrong}{很安静}。
(\colorbox{diffword}{Night} \colorbox{wrong}{is very quiet}.)}\\[0.2em]
\mbox{\textbf{FEA-SLT:} 
\colorbox{correct}{天黑了,}\colorbox{correct}{我害怕}\colorbox{correct}{。} 
(\colorbox{correct}{It is dark, I am afraid.})} \\

\midrule

\hbox{%
  \includegraphics[height=3\baselineskip, keepaspectratio, valign=t]{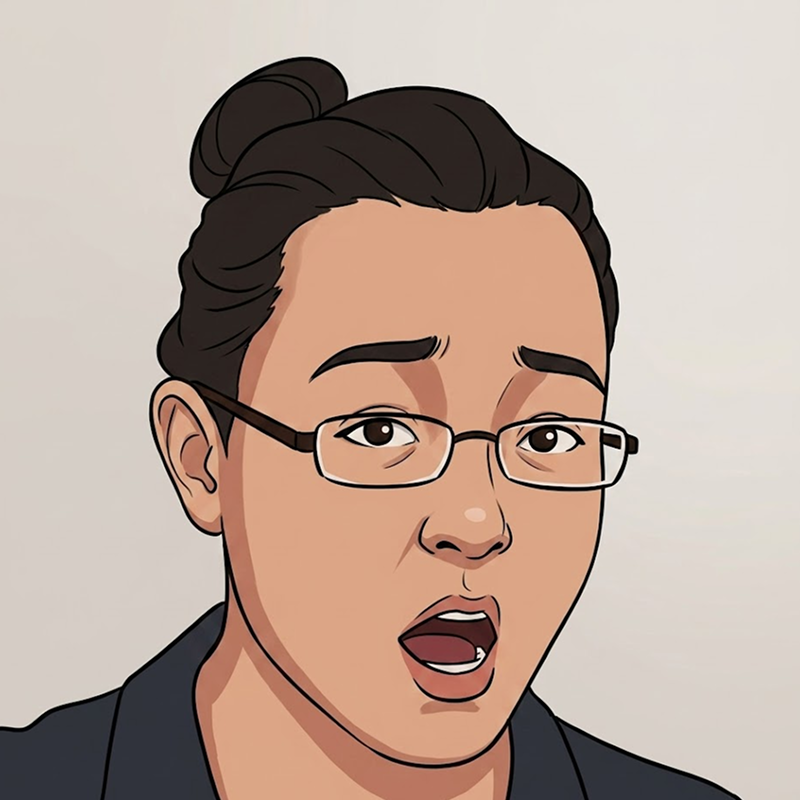}%
  \hspace{4pt}%
  \parbox[t]{\dimexpr\linewidth-0.18\linewidth-8pt\relax}{%
    \textbf{Reference:} 
    北京的名片是什么? 是人民大会堂。 
    (What is the name card of Beijing? It is the Great Hall of the People.)
  }
}
\textbf{SpaMo:} 
北京的\colorbox{wrong}{著名邮局}是人民大会堂。 
(\colorbox{wrong}{Famous post} \colorbox{wrong}{office in} Beijing is the Great Hall of the People.)\\[0.2em]
\textbf{FEA-SLT (w/o FE):} 
北京的\colorbox{correct}{名片} \colorbox{wrong}{(w/o ``是什么?’')} 是人民大会堂。 
(\colorbox{wrong}{(w/o ``What is the'')} \colorbox{correct}{name card} of Beijing is the Great Hall of the People.)\\[0.2em]
\textbf{FEA-SLT:} 
北京\colorbox{wrong}{总理}的\colorbox{correct}{名片}\colorbox{correct}{是什么?} 是人民大会堂。 
(\colorbox{correct}{What is the} \colorbox{wrong}{Premier's} \colorbox{correct}{name card of Beijing?} It is the Great Hall of the People.) \\
\bottomrule
\end{tabularx}
\caption{Qualitative examples from the CSL-Daily test set. Through the integration of facial-expression modeling, FEA-SLT better recovers facial-sensitive affective content and interrogative structure in these examples. \colorbox{wrong}{red} = incorrect, \colorbox{diffword}{yellow} = semantically correct but lexical variation, \colorbox{correct}{green} = fully correct. ($\cdots$) represents the English translation.}
\label{tab:csl_daily_qualitative}
\end{table}

\section{Conclusion}
\label{sec:conclusion}

We present FEA-SLT, a gloss-free end-to-end framework that explicitly incorporates facial expressions into SLT. 
Our approach employs a facial-expression encoding module based on pre-trained FER models to capture continuous facial dynamics, together with a facial-expression-aware fusion strategy that dynamically integrates manual and facial cues. 
Extensive experiments and ablation studies demonstrate that explicit facial-expression modeling consistently improves translation accuracy and semantic fidelity across multiple benchmarks. 
Qualitative and targeted subset analyses further suggest that facial expressions provide complementary facial-sensitive cues, particularly for utterances involving affective or interrogative semantics.
Overall, our findings highlight the importance of facial cues in gloss-free SLT and suggest that expression-sensitive facial modeling is a promising direction for future SLT research.

\section*{Limitations}
\label{sec:limitations}

Despite the promising empirical results, our proposed FEA-SLT framework exhibits several limitations.
First, the current framework models facial dynamics rather than the full range of NMS; mouthing (including proper-noun mouth patterns), head movements, and body posture remain outside its scope.
Second, FER-transferred representations do not explicitly disentangle affective expressions from grammatical facial markers.
Future work should study this distinction and use component-level masking or occlusion to measure the contribution of specific facial regions.
Third, all experiments use fixed seed 0, and the bootstrap analysis measures significance over fixed-model outputs rather than training variance.
Multi-seed runs and signer-stratified evaluation are needed to characterize optimization stability and generalization across individual expression styles.
Finally, facial-expression extraction remains sensitive to self-occlusion, illumination variation, and extreme head poses.
We report detailed failure statistics for both benchmarks in Appendix~\ref{sec:failure}.

\section*{Ethical Considerations}
\label{sec:ethics}

This research uses the publicly available PHOENIX14T and CSL-Daily datasets and follows their original licenses. Because facial-expression processing involves sensitive biometric information, raw facial images are discarded immediately after feature extraction; only de-identified latent representations are retained for academic research, and no derived artifacts are used outside this scope.
FER labels reflect affective categories that may vary across identities, cultures, and contexts, so transferred representations may conflate affective appearance with grammatical facial cues. They should not be interpreted as direct labels of a signer's emotional state or grammatical intent. Since our evaluation covers only DGS and CSL, future work should study broader sign languages, such as American Sign Language (ASL) and British Sign Language (BSL), and involve DHH community members to better align evaluation with community needs and cultural norms.

\section*{Acknowledgments}
We thank the anonymous reviewers for their valuable feedback.
This work was supported by Ningbo Yongjiang Talent Programme (2024A-399-G).


\bibliography{reference}

@inproceedings{li2025unisign,
  author    = {Zecheng Li and Wengang Zhou and Weichao Zhao and Kepeng Wu and Hezhen Hu and Houqiang Li},
  title     = {{UNI-SIGN}: Toward Unified Sign Language Understanding at Scale},
  booktitle = {Proceedings of the International Conference on Learning Representations (ICLR)},
  year      = {2025},
  url       = {https://openreview.net/forum?id=0Xt7uT04cQ}
}

@inproceedings{wong2024sign2gpt,
  author    = {Ryan Cameron Wong and Necati Cihan Camg{\"o}z and Richard Bowden},
  title     = {{Sign2GPT}: Leveraging Large Language Models for Gloss-Free Sign Language Translation},
  booktitle = {Proceedings of the International Conference on Learning Representations (ICLR)},
  year      = {2024},
  url       = {https://openreview.net/forum?id=LqaEEs3UxU}
}

@inproceedings{zhang2023sltunet,
    title={{SLTUNET}: A Simple Unified Model for Sign Language Translation},
    author={Biao Zhang and Mathias M{\"u}ller and Rico Sennrich},
    booktitle={Proceedings of the International Conference on Learning Representations (ICLR)},
    year={2023},
    url={https://openreview.net/forum?id=EBS4C77p_5S}
}

@inproceedings{hu2022lora,
    title={Lo{RA}: Low-Rank Adaptation of Large Language Models},
    author={Edward J Hu and Yelong Shen and Phillip Wallis and Zeyuan Allen-Zhu and Yuanzhi Li and Shean Wang and Lu Wang and Weizhu Chen},
    booktitle={Proceedings of the International Conference on Learning Representations (ICLR)},
    year={2022},
    url={https://openreview.net/forum?id=nZeVKeeFYf9}
}

@inproceedings{loshchilov2019decoupled,
  author    = {Ilya Loshchilov and Frank Hutter},
  title     = {Decoupled Weight Decay Regularization},
  booktitle = {Proceedings of the International Conference on Learning Representations (ICLR)},
  year      = {2019},
  url       = {https://openreview.net/forum?id=Bkg6RiCqY7}
}

@inproceedings{dosovitskiy2021image,
  title={An Image is Worth 16x16 Words: Transformers for Image Recognition at Scale},
  author={Dosovitskiy, Alexey and Beyer, Lucas and Kolesnikov, Alexander and Weissenborn, Dirk and Zhai, Xiaohua and Unterthiner, Thomas and  Dehghani, Mostafa and Minderer, Matthias and Heigold, Georg and Gelly, Sylvain and Uszkoreit, Jakob and Houlsby, Neil},
  booktitle={Proceedings of the 9th International Conference on Learning Representations (ICLR)},
  year={2021},
  url={https://openreview.net/forum?id=YicbFdNTTy}
}

@inproceedings{asasi2025beyondgloss,
  author    = {Sobhan Asasi and Mohamed Ilyes Lakhal and Ozge {Mercanoglu Sincan} and Richard Bowden},
  title     = {{Beyond Gloss}: A Hand-Centric Framework for Gloss-Free Sign Language Translation},
  booktitle = {Proceedings of the British Machine Vision Conference (BMVC)},
  year      = {2025},
  address   = {Sheffield, UK},
  publisher = {BMVA Press},
  url       = {https://bmva-archive.org.uk/bmvc/2025/assets/papers/Paper_626/paper.pdf}
}

@inproceedings{brown2020language,
 author = {Brown, Tom and Mann, Benjamin and Ryder, Nick and Subbiah, Melanie and Kaplan, Jared D and Dhariwal, Prafulla and Neelakantan, Arvind and Shyam, Pranav and Sastry, Girish and Askell, Amanda and Agarwal, Sandhini and Herbert-Voss, Ariel and Krueger, Gretchen and Henighan, Tom and Child, Rewon and Ramesh, Aditya and Ziegler, Daniel and Wu, Jeffrey and Winter, Clemens and Hesse, Chris and Chen, Mark and Sigler, Eric and Litwin, Mateusz and Gray, Scott and Chess, Benjamin and Clark, Jack and Berner, Christopher and McCandlish, Sam and Radford, Alec and Sutskever, Ilya and Amodei, Dario},
 booktitle = {Advances in Neural Information Processing Systems},
 editor = {H. Larochelle and M. Ranzato and R. Hadsell and M.F. Balcan and H. Lin},
 pages = {1877--1901},
 publisher = {Curran Associates, Inc.},
 title = {Language Models are Few-Shot Learners},
 url = {https://proceedings.neurips.cc/paper_files/paper/2020/file/1457c0d6bfcb4967418bfb8ac142f64a-Paper.pdf},
 volume = {33},
 year = {2020}
}

@inproceedings{chen2022ts-slt,
 author = {Chen, Yutong and Zuo, Ronglai and Wei, Fangyun and Wu, Yu and LIU, Shujie and Mak, Brian},
 booktitle = {Advances in Neural Information Processing Systems},
 editor = {S. Koyejo and S. Mohamed and A. Agarwal and D. Belgrave and K. Cho and A. Oh},
 pages = {17043--17056},
 publisher = {Curran Associates, Inc.},
 title = {Two-Stream Network for Sign Language Recognition and Translation},
 url = {https://proceedings.neurips.cc/paper_files/paper/2022/file/6cd3ac24cdb789beeaa9f7145670fcae-Paper-Conference.pdf},
 volume = {35},
 year = {2022}
}

@inproceedings{tong2022videomae,
    author = {Tong, Zhan and Song, Yibing and Wang, Jue and Wang, Limin},
    title = {{VideoMAE}: Masked Autoencoders Are Data-Efficient Learners for Self-Supervised Video Pre-Training},
    booktitle = {Advances in Neural Information Processing Systems},
    volume = {35},
    pages = {10078--10093},
    year = {2022},
    publisher = {Curran Associates, Inc.},
    url = {https://proceedings.neurips.cc/paper_files/paper/2022/file/416f9cb3276121c42eebb86352a4354a-Paper-Conference.pdf},
    doi = {10.52202/068431-0732}
}

@inproceedings{li2020tspnet,
	title        = {TSPNet: Hierarchical Feature Learning via Temporal Semantic Pyramid for Sign Language Translation},
	author       = {Li, Dongxu and Xu, Chenchen and Yu, Xin and Zhang, Kaihao and Swift, Benjamin and Suominen, Hanna and Li, Hongdong},
	year         = 2020,
	booktitle    = {Advances in Neural Information Processing Systems},
	volume       = 33,
	url          = {https://proceedings.neurips.cc/paper_files/paper/2020/hash/8c00dee24c9878fea090ed070b44f1ab-Abstract.html}
}

@inproceedings{szegedy2016rethinking,
  author    = {Christian Szegedy and Vincent Vanhoucke and Sergey Ioffe and Jon Shlens and Zbigniew Wojna},
  title     = {Rethinking the inception architecture for computer vision},
  booktitle = {Proceedings of the IEEE/CVF Conference on Computer Vision and Pattern Recognition (CVPR)},
  year      = {2016},
  pages     = {2818--2826},
  doi       = {10.1109/CVPR.2016.308}
}

@inproceedings{camgoz2018neural,
  author    = {Necati Cihan Camg{\"o}z and Simon Hadfield and Oscar Koller and Hermann Ney and Richard Bowden},
  title     = {Neural Sign Language Translation},
  booktitle = {Proceedings of the IEEE/CVF Conference on Computer Vision and Pattern Recognition (CVPR)},
  year      = {2018},
  pages     = {7784--7793},
  doi       = {10.1109/CVPR.2018.00812}
}

@inproceedings{gong2024llms,
  author    = {Jia Gong and Lin Geng Foo and Yixuan He and Hossein Rahmani and Jun Liu},
  title     = {LLMs Are Good Sign Language Translators},
  booktitle = {Proceedings of the IEEE/CVF Conference on Computer Vision and Pattern Recognition (CVPR)},
  year      = {2024},
  pages     = {18362--18372},
  doi       = {10.1109/CVPR52733.2024.01738}
}

@inproceedings{jang2025lost,
    title={Lost in Translation, Found in Context: Sign Language Translation with Contextual Cues},
    author={Jang, Youngjoon and Raajesh, Haran and Momeni, Liliane and Varol, G{\"u}l and Zisserman, Andrew},
    booktitle={Proceedings of the IEEE/CVF Conference on Computer Vision and Pattern Recognition (CVPR)},
    year={2025},
    pages     = {8742--8752},
    doi       = {10.1109/CVPR52734.2025.00817}
  }

@InProceedings{hu2023continuous,
    author    = {Hu, Lianyu and Gao, Liqing and Liu, Zekang and Feng, Wei},
    title     = {Continuous Sign Language Recognition With Correlation Network},
    booktitle = {Proceedings of the IEEE/CVF Conference on Computer Vision and Pattern Recognition (CVPR)},
    month     = {June},
    year      = {2023},
    pages     = {2529-2539},
    url       = {https://arxiv.org/abs/2303.03202}
}

@InProceedings{liu2024improved,
    author    = {Liu, Haotian and Li, Chunyuan and Li, Yuheng and Lee, Yong Jae},
    title     = {Improved Baselines with Visual Instruction Tuning},
    booktitle = {Proceedings of the IEEE/CVF Conference on Computer Vision and Pattern Recognition (CVPR)},
    month     = {June},
    year      = {2024},
    pages     = {26296-26306},
    doi       = {10.1109/CVPR52733.2024.02484}
}

@inproceedings{yin2023gloss,
  title={Gloss attention for gloss-free sign language translation},
  author={Yin, Aoxiong and Zhong, Tianyun and Tang, Li and Jin, Weike and Jin, Tao and Zhao, Zhou},
  booktitle={Proceedings of the IEEE/CVF Conference on Computer Vision and Pattern Recognition (CVPR)},
  pages={2551--2562},
  year={2023},
  doi={10.1109/CVPR52729.2023.00251},
  url={https://openaccess.thecvf.com/content/CVPR2023/html/Yin_Gloss_Attention_for_Gloss-Free_Sign_Language_Translation_CVPR_2023_paper.html}
}

@inproceedings{zhou2021improving,
  title={Improving sign language translation with monolingual data by sign back-translation},
  author={Zhou, Hao and Zhou, Wengang and Qi, Weizhen and Pu, Junfu and Li, Houqiang},
  booktitle={Proceedings of the IEEE/CVF Conference on Computer Vision and Pattern Recognition (CVPR)},
  pages={1316--1325},
  year={2021},
  doi={10.1109/CVPR46437.2021.00137}
}

@inproceedings{xue2021mt5,
    title = "m{T}5: A Massively Multilingual Pre-trained Text-to-Text Transformer",
    author = "Xue, Linting  and
      Constant, Noah  and
      Roberts, Adam  and
      Kale, Mihir  and
      Al-Rfou, Rami  and
      Siddhant, Aditya  and
      Barua, Aditya  and
      Raffel, Colin",
    editor = "Toutanova, Kristina  and
      Rumshisky, Anna  and
      Zettlemoyer, Luke  and
      Hakkani-Tur, Dilek  and
      Beltagy, Iz  and
      Bethard, Steven  and
      Cotterell, Ryan  and
      Chakraborty, Tanmoy  and
      Zhou, Yichao",
    booktitle = "Proceedings of the North American Chapter of the Association for Computational Linguistics: Human Language Technologies (NAACL-HLT)",
    month = jun,
    year = "2021",
    address = "Online",
    publisher = "Association for Computational Linguistics",
    url = "https://aclanthology.org/2021.naacl-main.41/",
    doi = "10.18653/v1/2021.naacl-main.41",
    pages = "483--498",
}

@inproceedings{devlin2019bert,
  author    = {Jacob Devlin and Ming-Wei Chang and Kenton Lee and Kristina Toutanova},
  title     = {{BERT}: Pre-training of Deep Bidirectional Transformers for Language Understanding},
  booktitle = {Proceedings of the North American Chapter of the Association for Computational Linguistics: Human Language Technologies (NAACL-HLT)},
  year      = {2019},
  pages     = {4171--4186},
  doi       = {10.18653/v1/N19-1423},
  url       = {https://aclanthology.org/N19-1423/}
}

@inproceedings{hwang2025spamo,
  author    = {Eui Jun Hwang and Sukmin Cho and Junmyeong Lee and Jong C. Park},
  title     = {An Efficient Gloss-Free Sign Language Translation Using Spatial Configurations and Motion Dynamics with LLMs},
  booktitle = {Proceedings of the North American Chapter of the Association for Computational Linguistics: Human Language Technologies (NAACL-HLT)},
  year      = {2025},
  pages     = {3901--3920},
  address   = {Albuquerque, New Mexico, USA},
  publisher = {Association for Computational Linguistics},
  url       = {https://aclanthology.org/2025.naacl-long.197/},
  doi       = {10.18653/v1/2025.naacl-long.197}
}

@misc{bai2018tcn,
      title={An Empirical Evaluation of Generic Convolutional and Recurrent Networks for Sequence Modeling}, 
      author={Shaojie Bai and J. Zico Kolter and Vladlen Koltun},
      year={2018},
      eprint={1803.01271},
      archivePrefix={arXiv},
      primaryClass={cs.LG},
      url={https://arxiv.org/abs/1803.01271}, 
}

@misc{chung2022flant5,
      title={Scaling Instruction-Finetuned Language Models}, 
      author={Hyung Won Chung and Le Hou and Shayne Longpre and Barret Zoph and Yi Tay and William Fedus and Yunxuan Li and Xuezhi Wang and Mostafa Dehghani and Siddhartha Brahma and Albert Webson and Shixiang Shane Gu and Zhuyun Dai and Mirac Suzgun and Xinyun Chen and Aakanksha Chowdhery and Alex Castro-Ros and Marie Pellat and Kevin Robinson and Dasha Valter and Sharan Narang and Gaurav Mishra and Adams Yu and Vincent Zhao and Yanping Huang and Andrew Dai and Hongkun Yu and Slav Petrov and Ed H. Chi and Jeff Dean and Jacob Devlin and Adam Roberts and Denny Zhou and Quoc V. Le and Jason Wei},
      year={2022},
      eprint={2210.11416},
      archivePrefix={arXiv},
      primaryClass={cs.LG},
      url={https://arxiv.org/abs/2210.11416}, 
}

@inproceedings{chen2022simple,
  title={A simple multi-modality transfer learning baseline for sign language translation},
  author={Chen, Yutong and Wei, Fangyun and Sun, Xiao and Wu, Zhirong and Lin, Stephen},
  booktitle={Proceedings of the IEEE/CVF Conference on Computer Vision and Pattern Recognition (CVPR)},
  pages={5120--5130},
  year={2022},
  doi={10.1109/CVPR52688.2022.00506},
  url={https://openaccess.thecvf.com/content/CVPR2022/html/Chen_A_Simple_Multi-Modality_Transfer_Learning_Baseline_for_Sign_Language_Translation_CVPR_2022_paper.html}
}

@inproceedings{camgoz2020sign,
  author    = {Necati Cihan Camg{\"o}z and Oscar Koller and Simon Hadfield and Richard Bowden},
  title     = {Sign Language Transformers: Joint End-to-End Sign Language Recognition and Translation},
  booktitle   = {Proceedings of the IEEE/CVF Conference on Computer Vision and Pattern Recognition (CVPR)},
  year      = {2020},
  pages     = {10023--10033},
  doi       = {10.1109/CVPR42600.2020.01004}
}

@inproceedings{kim2025mmslt,
  author    = {Jungeun Kim and Hyeongwoo Jeon and Jongseong Bae and Ha Young Kim},
  title     = {Leveraging the Power of MLLMs for Gloss-Free Sign Language Translation},
  booktitle = {Proceedings of the IEEE/CVF International Conference on Computer Vision (ICCV)},
  year      = {2025},
  pages     = {21048--21058},
  url       = {https://openaccess.thecvf.com/content/ICCV2025/html/Kim_Leveraging_the_Power_of_MLLMs_for_Gloss-Free_Sign_Language_Translation_ICCV_2025_paper.html}
}

@inproceedings{zhou2023glofe,
  author    = {Benjia Zhou and Zhigang Chen and Albert Clap{\'e}s and Jun Wan and Yanyan Liang and Sergio Escalera and Zhen Lei and Du Zhang},
  title     = {Gloss-Free Sign Language Translation: Improving from Visual-Language Pretraining},
  booktitle = {Proceedings of the IEEE/CVF International Conference on Computer Vision (ICCV)},
  year      = {2023},
  pages     = {20871--20881},
  url       = {https://openaccess.thecvf.com/content/ICCV2023/html/Zhou_Gloss-Free_Sign_Language_Translation_Improving_from_Visual-Language_Pretraining_ICCV_2023_paper.html}
}

@inproceedings{jiao2024vap,
  author    = {Peiqi Jiao and Yuecong Min and Xilin Chen},
  title     = {Visual Alignment Pre-training for Sign Language Translation},
  booktitle = {Proceedings of the European Conference on Computer Vision (ECCV)},
  year      = {2024},
  pages     = {349--367},
  doi       = {10.1007/978-3-031-72946-1_20}
}

@inproceedings{jiao2023cosign,
  author       = {Peiqi Jiao and
                  Yuecong Min and
                  Yanan Li and
                  Xiaotao Wang and
                  Lei Lei and
                  Xilin Chen},
  title        = {CoSign: Exploring Co-occurrence Signals in Skeleton-based Continuous
                  Sign Language Recognition},
  booktitle    = {Proceedings of the IEEE/CVF International Conference on Computer Vision (ICCV)},
  pages        = {20619--20629},
  publisher    = {{IEEE}},
  year         = {2023},
  url          = {https://doi.org/10.1109/ICCV51070.2023.01890},
  doi          = {10.1109/ICCV51070.2023.01890},
}

@inproceedings{deng2019retinaface,
  title={RetinaFace: Single-Shot Multi-Level Face Localisation in the Wild},
  author={Deng, Jiankang and Guo, Jia and Ververas, Evangelos and Kotsia, Irene and Zafeiriou, Stefanos},
  booktitle={Proceedings of the IEEE/CVF Conference on Computer Vision and Pattern Recognition (CVPR)},
  pages={5203--5212},
  year={2020},
  doi={10.1109/CVPR42600.2020.00525}
}

@inproceedings{chen2024fla,
    title = "Factorized Learning Assisted with Large Language Model for Gloss-free Sign Language Translation",
    author = "Chen, Zhigang  and
      Zhou, Benjia  and
      Li, Jun  and
      Wan, Jun  and
      Lei, Zhen  and
      Jiang, Ning  and
      Lu, Quan  and
      Zhao, Guoqing",
    editor = "Calzolari, Nicoletta  and
      Kan, Min-Yen  and
      Hoste, Veronique  and
      Lenci, Alessandro  and
      Sakti, Sakriani  and
      Xue, Nianwen",
    booktitle = "Proceedings of the 2024 Joint International Conference on Computational Linguistics, Language Resources and Evaluation (LREC-COLING 2024)",
    month = may,
    year = "2024",
    address = "Torino, Italia",
    publisher = "ELRA and ICCL",
    url = "https://aclanthology.org/2024.lrec-main.620/",
    pages = "7071--7081"
}

@inproceedings{sharma2024facial,
    title = "Facial Expressions for Sign Language Synthesis using {FACSH}uman and {AZ}ee",
    author = "Sharma, Paritosh  and
      Challant, Camille  and
      Filhol, Michael",
    editor = "Efthimiou, Eleni  and
      Fotinea, Stavroula-Evita  and
      Hanke, Thomas  and
      Hochgesang, Julie A.  and
      Mesch, Johanna  and
      Schulder, Marc",
    booktitle = "Proceedings of the LREC-COLING 2024 11th Workshop on the Representation and Processing of Sign Languages: Evaluation of Sign Language Resources",
    month = may,
    year = "2024",
    address = "Torino, Italia",
    publisher = "ELRA and ICCL",
    url = "https://aclanthology.org/2024.signlang-1.39/",
    pages = "354--360"
}

@misc{chua2025emosign,
      title={{EmoSign}: A Multimodal Dataset for Understanding Emotions in American Sign Language}, 
      author={Phoebe Chua and Cathy Mengying Fang and Takehiko Ohkawa and Raja Kushalnagar and Suranga Nanayakkara and Pattie Maes},
      year={2025},
      eprint={2505.17090},
      archivePrefix={arXiv},
      primaryClass={cs.CV},
      url={https://arxiv.org/abs/2505.17090}, 
}

@article{elliott2013facial,
  author    = {Eeva A. Elliott and Arthur M. Jacobs},
  title     = {Facial Expressions, Emotions, and Sign Languages},
  journal   = {Frontiers in Psychology},
  year      = {2013},
  volume    = {4},
  pages     = {115},
  doi       = {10.3389/fpsyg.2013.00115},
  url       = {https://www.frontiersin.org/journals/psychology/articles/10.3389/fpsyg.2013.00115/full}
}

@inproceedings{fu2023conslt,
  author = {Fu, Biao and Ye, Peigen and Zhang, Liang and Yu, Pei and Hu, Cong and Chen, Yidong and Shi, Xiaodong},
  title = {A Token-level Contrastive Framework for Sign Language Translation},
  booktitle = {Proceedings of the IEEE International Conference on Acoustics, Speech and Signal Processing (ICASSP)},
  year = {2023},
  doi = {10.1109/ICASSP49357.2023.10095466},
  url = {https://arxiv.org/abs/2204.04916}
}

@misc{trpakov2023vitface,
  title={ViT Face Expression Recognition Model},
  author={Trpakov, Stanislav},
  year={2023},
  howpublished={\url{https://huggingface.co/trpakov/vit-face-expression}},
  url={https://huggingface.co/trpakov/vit-face-expression},
  note={Accessed: 2023-12-13}
}

@inproceedings{radford2021learning,
  title={Learning transferable visual models from natural language supervision},
  author={Radford, Alec and Kim, Jong Wook and Hallacy, Chris and Ramesh, Aditya and Goh, Gabriel and Agarwal, Sandhini and Sastry, Girish and Askell, Amanda and Mishkin, Pamela and Clark, Jack and others},
  booktitle={Proceedings of the International Conference on Machine Learning (ICML)},
  pages={8748--8763},
  year={2021},
  organization={PMLR},
  url={https://proceedings.mlr.press/v139/radford21a.html}
}

@article{goodfellow2013challenges,
  title={Challenges in representation learning: A report on three machine learning contests},
  author={Goodfellow, Ian J and Erhan, Dumitru and Carrier, Pierre Luc and Courville, Aaron and Mirza, Mehdi and Hamner, Ben and Cukierski, Will and Tang, Yichuan and Thaler, David and Lee, Dong-Hyun and others},
  journal={Neural Networks},
  volume={64},
  pages={59--63},
  year={2015},
  doi={10.1016/j.neunet.2014.09.005},
  url={https://arxiv.org/abs/1307.0414}
}

@inproceedings{lin2004rouge,
    title = "{ROUGE}: A Package for Automatic Evaluation of Summaries",
    author = "Lin, Chin-Yew",
    booktitle = "Proceedings of the Workshop on Text Summarization Branches Out",
    month = jul,
    year = "2004",
    address = "Barcelona, Spain",
    publisher = "Association for Computational Linguistics",
    url = "https://aclanthology.org/W04-1013/",
    pages = "74--81"
}

@inproceedings{lin2023glofe,
  author    = {Kezhou Lin and Xiaohan Wang and Linchao Zhu and Ke Sun and Bang Zhang and Yi Yang},
  title     = {Gloss-Free End-to-End Sign Language Translation},
  booktitle = {Proceedings of the 61st Annual Meeting of the Association for Computational Linguistics (Volume 1: Long Papers)},
  year      = {2023},
  pages     = {12904--12916},
  address   = {Toronto, Canada},
  publisher = {Association for Computational Linguistics},
  url       = {https://aclanthology.org/2023.acl-long.722/}
}

@inproceedings{rust2024towards,
    title = "Towards Privacy-Aware Sign Language Translation at Scale",
    author = "Rust, Phillip  and
      Shi, Bowen  and
      Wang, Skyler  and
      Camgoz, Necati Cihan  and
      Maillard, Jean",
    editor = "Ku, Lun-Wei  and
      Martins, Andre  and
      Srikumar, Vivek",
    booktitle = "Proceedings of the 62nd Annual Meeting of the Association for Computational Linguistics (Volume 1: Long Papers)",
    month = aug,
    year = "2024",
    address = "Bangkok, Thailand",
    publisher = "Association for Computational Linguistics",
    url = "https://aclanthology.org/2024.acl-long.467/",
    doi = "10.18653/v1/2024.acl-long.467",
    pages = "8624--8641"
}

@article{liu2020multilingual,
    title = "Multilingual Denoising Pre-training for Neural Machine Translation",
    author = "Liu, Yinhan  and
      Gu, Jiatao  and
      Goyal, Naman  and
      Li, Xian  and
      Edunov, Sergey  and
      Ghazvininejad, Marjan  and
      Lewis, Mike  and
      Zettlemoyer, Luke",
    editor = "Johnson, Mark  and
      Roark, Brian  and
      Nenkova, Ani",
    journal = "Transactions of the Association for Computational Linguistics (TACL)",
    volume = "8",
    year = "2020",
    address = "Cambridge, MA",
    publisher = "MIT Press",
    url = "https://aclanthology.org/2020.tacl-1.47/",
    doi = "10.1162/tacl_a_00343",
    pages = "726--742",
}

@misc{tang2020multilingual,
    title = {Multilingual Translation with Extensible Multilingual Pretraining and Finetuning},
    author = {Tang, Yuqing and Tran, Chau and Li, Xian and Chen, Peng-Jen and Goyal, Naman and Chaudhary, Vishrav and Gu, Jiatao and Fan, Angela},
    year = {2020},
    eprint = {2008.00401},
    archivePrefix = {arXiv},
    primaryClass = {cs.CL},
    doi = {10.48550/arXiv.2008.00401},
    url = {https://arxiv.org/abs/2008.00401}
}

@inproceedings{pu2021learning,
  title = {Learning compact metrics for MT},
  author = {Pu, Amy and Chung, Hyung Won and Parikh, Ankur P. and Gehrmann, Sebastian and Sellam, Thibault},
  booktitle = {Proceedings of the 2021 Conference on Empirical Methods in Natural Language Processing (EMNLP)},
  year = {2021},
  pages = {751--762},
  doi = {10.18653/v1/2021.emnlp-main.58},
  url = {https://aclanthology.org/2021.emnlp-main.58/}
}

@inproceedings{luong2015effective,
  author    = {Luong, Thang and Pham, Hieu and Manning, Christopher D.},
  title     = {Effective Approaches to Attention-Based Neural Machine Translation},
  booktitle = {Proceedings of the 2015 Conference on Empirical Methods in Natural Language Processing (EMNLP)},
  year      = {2015},
  pages     = {1412--1421},
  doi       = {10.18653/v1/D15-1166},
  url       = {https://aclanthology.org/D15-1166/}
}

@inproceedings{shi2024when,
      title={When Do We Not Need Larger Vision Models?},
      author={Shi, Baifeng and Wu, Ziyang and Mao, Maolin and Wang, Xin and Darrell, Trevor},
      booktitle={Proceedings of the European Conference on Computer Vision (ECCV)},
      pages={444--462},
      year={2024},
      doi={10.1007/978-3-031-73242-3_25},
      url={https://arxiv.org/abs/2403.13043},
}

@misc{openai2024gpt4technicalreport,
      title={GPT-4 Technical Report},
      author={{OpenAI}},
      year={2023},
      eprint={2303.08774},
      archivePrefix={arXiv},
      primaryClass={cs.CL},
      url={https://arxiv.org/abs/2303.08774},
}

@inproceedings{jin2022prior,
    title = "Prior Knowledge and Memory Enriched Transformer for Sign Language Translation",
    author = "Jin, Tao  and
      Zhao, Zhou  and
      Zhang, Meng  and
      Zeng, Xingshan",
    editor = "Muresan, Smaranda  and
      Nakov, Preslav  and
      Villavicencio, Aline",
    booktitle = "Findings of the Association for Computational Linguistics: ACL 2022",
    month = may,
    year = "2022",
    address = "Dublin, Ireland",
    publisher = "Association for Computational Linguistics",
    url = "https://aclanthology.org/2022.findings-acl.297/",
    doi = "10.18653/v1/2022.findings-acl.297",
    pages = "3766--3775"
}

@inproceedings{sellam2020bleurt,
    title = "{BLEURT}: Learning Robust Metrics for Text Generation",
    author = "Sellam, Thibault  and
      Das, Dipanjan  and
      Parikh, Ankur",
    editor = "Jurafsky, Dan  and
      Chai, Joyce  and
      Schluter, Natalie  and
      Tetreault, Joel",
    booktitle = "Proceedings of the 58th Annual Meeting of the Association for Computational Linguistics",
    month = jul,
    year = "2020",
    address = "Online",
    publisher = "Association for Computational Linguistics",
    url = "https://aclanthology.org/2020.acl-main.704/",
    doi = "10.18653/v1/2020.acl-main.704",
    pages = "7881--7892"
}

@inproceedings{hamidullah2024sem,
    title = "Sign Language Translation with Sentence Embedding Supervision",
    author = "Hamidullah, Yasser  and
      van Genabith, Josef  and
      Espa{\~n}a-Bonet, Cristina",
    editor = "Ku, Lun-Wei  and
      Martins, Andre  and
      Srikumar, Vivek",
    booktitle = "Proceedings of the 62nd Annual Meeting of the Association for Computational Linguistics (Volume 2: Short Papers)",
    month = aug,
    year = "2024",
    address = "Bangkok, Thailand",
    publisher = "Association for Computational Linguistics",
    url = "https://aclanthology.org/2024.acl-short.40/",
    doi = "10.18653/v1/2024.acl-short.40",
    pages = "425--434"
}

@inproceedings{viegas2023including,
    title = "Including Facial Expressions in Contextual Embeddings for Sign Language Generation",
    author = "Viegas, Carla  and
      Inan, Mert  and
      Quandt, Lorna  and
      Alikhani, Malihe",
    editor = "Palmer, Alexis  and
      Camacho-collados, Jose",
    booktitle = "Proceedings of the 12th Joint Conference on Lexical and Computational Semantics (*SEM 2023)",
    month = jul,
    year = "2023",
    address = "Toronto, Canada",
    publisher = "Association for Computational Linguistics",
    url = "https://aclanthology.org/2023.starsem-1.1/",
    doi = "10.18653/v1/2023.starsem-1.1",
    pages = "1--10"
}

@misc{chua2025perspectives,
      title={Perspectives on Capturing Emotional Expressiveness in Sign Language}, 
      author={Phoebe Chua and Cathy Mengying Fang and Yasith Samaradivakara and Pattie Maes and Suranga Nanayakkara},
      year={2025},
      eprint={2505.08072},
      archivePrefix={arXiv},
      primaryClass={cs.HC},
      url={https://arxiv.org/abs/2505.08072}, 
}

@inproceedings{papineni2002bleu,
    title = "{B}leu: a Method for Automatic Evaluation of Machine Translation",
    author = "Papineni, Kishore  and
      Roukos, Salim  and
      Ward, Todd  and
      Zhu, Wei-Jing",
    editor = "Isabelle, Pierre  and
      Charniak, Eugene  and
      Lin, Dekang",
    booktitle = "Proceedings of the 40th Annual Meeting of the Association for Computational Linguistics",
    month = jul,
    year = "2002",
    address = "Philadelphia, Pennsylvania, USA",
    publisher = "Association for Computational Linguistics",
    url = "https://aclanthology.org/P02-1040/",
    doi = "10.3115/1073083.1073135",
    pages = "311--318"
}

@book{pfau2012sign,
  editor    = {Roland Pfau and Markus Steinbach and Bencie Woll},
  title     = {Sign Language: An International Handbook},
  publisher = {De Gruyter Mouton},
  year      = {2012},
  doi       = {10.1515/9783110261325}
}

@misc{rastgoo2022all,
      title={All You Need In Sign Language Production}, 
      author={Razieh Rastgoo and Kourosh Kiani and Sergio Escalera and Vassilis Athitsos and Mohammad Sabokrou},
      year={2022},
      eprint={2201.01609},
      archivePrefix={arXiv},
      primaryClass={cs.CV},
      url={https://arxiv.org/abs/2201.01609}, 
}

@article{reilly1992affective,
 ISSN = {03021475, 15336263},
 URL = {http://www.jstor.org/stable/26204636},
 author = {Judy S. Reilly and Marina L. McIntire and Howie Seago},
 journal = {Sign Language Studies},
 volume = {75},
 pages = {113--128},
 publisher = {Gallaudet University Press},
 title = {Affective Prosody in American Sign Language},
 urldate = {2026-01-05},
 doi = {10.1353/sls.1992.0035},
 year = {1992}
}

@article{zhao2022conditional,
  author    = {Jian Zhao and Weizhen Qi and Wengang Zhou and Nan Duan and Ming Zhou and Houqiang Li},
  title     = {Conditional Sentence Generation and Cross-Modal Reranking for Sign Language Translation},
  journal   = {IEEE Transactions on Multimedia},
  year      = {2022},
  volume    = {24},
  pages     = {2662--2672},
  doi       = {10.1109/TMM.2021.3087006},
}

@inproceedings{tan2025mlslt,
    title = {Multilingual Gloss-free Sign Language Translation: Towards Building a Sign Language Foundation Model},
    author = {Tan, Sihan and Miyazaki, Taro and Nakadai, Kazuhiro},
    booktitle = {Proceedings of the 63rd Annual Meeting of the Association for Computational Linguistics (Volume 2: Short Papers)},
    month = jul,
    year = {2025},
    address = {Vienna, Austria},
    publisher = {Association for Computational Linguistics},
    pages = {553--561},
    doi = {10.18653/v1/2025.acl-short.43},
    url = {https://aclanthology.org/2025.acl-short.43/}
}

@inproceedings{gueuwou2025shubert,
    title = {SHuBERT: Self-Supervised Sign Language Representation Learning via Multi-Stream Cluster Prediction},
    author = "Gueuwou, Shester and Du, Xiaodan and Shakhnarovich, Greg and Livescu, Karen and Liu, Alexander H.",
    booktitle = "Proceedings of the 63rd Annual Meeting of the Association for Computational Linguistics (Volume 1: Long Papers)",
    month = jul,
    year = {2025},
    address = "Vienna, Austria",
    publisher = "Association for Computational Linguistics",
    pages = {28792--28810},
    doi = {10.18653/v1/2025.acl-long.1397},
    url = {https://aclanthology.org/2025.acl-long.1397/}
}

@inproceedings{gueuwou2025signmusketeers,
      title = "{S}ign{M}usketeers: An Efficient Multi-Stream Approach for Sign Language Translation at Scale",
      author = "Gueuwou, Shester  and
        Du, Xiaodan  and
        Shakhnarovich, Greg  and
        Livescu, Karen",
      editor = "Che, Wanxiang  and
        Nabende, Joyce  and
        Shutova, Ekaterina  and
        Pilehvar, Mohammad Taher",
      booktitle = "Findings of the Association for Computational Linguistics: ACL 2025",
      month = jul,
      year = "2025",
      address = "Vienna, Austria",
      publisher = "Association for Computational Linguistics",
      url = "https://aclanthology.org/2025.findings-acl.1157/",
      doi = "10.18653/v1/2025.findings-acl.1157",
      pages = "22506--22521",
      ISBN = "979-8-89176-256-5",
  }

@inproceedings{post2018call,
  title = {A Call for Clarity in Reporting {BLEU} Scores},
  author = {Post, Matt},
  booktitle = {Proceedings of the Conference on Machine Translation (WMT): Research Papers},
  month = oct,
  year = "2018",
  address = "Belgium, Brussels",
  publisher = "Association for Computational Linguistics",
  url = "https://www.aclweb.org/anthology/W18-6319",
  pages = "186--191",
}

@inproceedings{hamidullah2025sonar,
    title = "{SONAR}-{SLT}: Multilingual Sign Language Translation via Language-Agnostic Sentence Embedding Supervision",
    author = "Hamidullah, Yasser  and
      Yazdani, Shakib  and
      Oguz, Cennet  and
      Van Genabith, Josef  and
      Espa{\~n}a-Bonet, Cristina",
    editor = "Haddow, Barry  and
      Kocmi, Tom  and
      Koehn, Philipp  and
      Monz, Christof",
    booktitle = "Proceedings of the Conference on Machine Translation (WMT)",
    month = nov,
    year = "2025",
    address = "Suzhou, China",
    publisher = "Association for Computational Linguistics",
    url = "https://aclanthology.org/2025.wmt-1.18/",
    doi = "10.18653/v1/2025.wmt-1.18",
    pages = "301--313",
    ISBN = "979-8-89176-341-8"
}

@article{nmfs2021hu,
    author = {Hu, Hezhen and Zhou, Wengang and Pu, Junfu and Li, Houqiang},
    title = {Global-Local Enhancement Network for NMF-Aware Sign Language Recognition},
    year = {2021},
    issue_date = {August 2021},
    publisher = {Association for Computing Machinery (ACM)},
    address = {New York, NY, USA},
    volume = {17},
    number = {3},
    issn = {1551-6857},
    url = {https://doi.org/10.1145/3436754},
    doi = {10.1145/3436754},
    journal = {ACM Trans. Multimedia Comput. Commun. Appl.},
    month = jul,
    articleno = {80},
    numpages = {19}
}

@inproceedings{pytorch2019paszke,
    author    = {Paszke, Adam and Gross, Sam and Massa, Francisco and Lerer, Adam and Bradbury, James and Chanan, Gregory and Killeen, Trevor and Lin, Zeming and Gimelshein, Natalia and Antiga, Luca and Desmaison, Alban and Kopf, Andreas and Yang, Edward and DeVito, Zachary and Raison, Martin and Tejani, Alykhan and Chilamkurthy, Sasank and Steiner, Benoit and Fang, Lu and Bai, Junjie and Chintala, Soumith},
    title     = {PyTorch: An Imperative Style, High-Performance Deep Learning Library},
    booktitle = {Advances in Neural Information Processing Systems},
    volume    = {32},
    pages     = {8024--8035},
    year      = {2019},
    url       = {https://proceedings.neurips.cc/paper/2019/hash/bdbca288fee7f92f2bfa9f7012727740-Abstract.html}
}

@incollection{pfau2010nonmanuals,
    author = {Pfau, Roland and Quer, Josep},
    year = {2010},
    pages = {381--402},
    title = {Nonmanuals: Their prosodic and grammatical roles},
    booktitle = {Sign Languages},
    editor = {Brentari, Diane},
    publisher = {Cambridge University Press},
    isbn = {9780511712203},
    doi = {10.1017/CBO9780511712203.018}
}

@article{brentari2002prosody,
  title={Prosody on the hands and face: Evidence from American Sign Language},
  author={Brentari, Diane and Crossley, Laurinda},
  journal={Sign Language \& Linguistics},
  volume={5},
  pages={105--130},
  year={2002},
  publisher={John Benjamins Publishing Company},
  doi={10.1075/sll.5.2.03bre},
  url={https://www.jbe-platform.com/content/journals/10.1075/sll.5.2.03bre}
}

@inproceedings{jiang2026think,
  title     = {Think in Latent Thoughts: A New Paradigm for Gloss-Free Sign Language Translation},
  author    = {Jiang, Yiyang and Zhang, Li and Wei, Xiao-Yong and Qing, Li},
  booktitle = {Proceedings of the 64th Annual Meeting of the Association for Computational Linguistics (Volume 1: Long Papers)},
  month     = jul,
  year      = {2026},
  address   = {San Diego, California, United States},
  publisher = {Association for Computational Linguistics},
  pages     = {9993--10012},
  doi       = {10.18653/v1/2026.acl-long.454},
  url       = {https://aclanthology.org/2026.acl-long.454/}
}

@article{mcinnes2018umap,
  doi = {10.21105/joss.00861},
  url = {https://doi.org/10.21105/joss.00861},
  year = {2018},
  publisher = {The Open Journal},
  volume = {3},
  number = {29},
  pages = {861},
  author = {McInnes, Leland and Healy, John and Saul, Nathaniel and Großberger, Lukas},
  title = {UMAP: Uniform Manifold Approximation and Projection},
  journal = {Journal of Open Source Software}
}

@book{ekman1978facs,
  title={Facial Action Coding System: A Technique for the Measurement of Facial Movement},
  author={Ekman, Paul and Friesen, Wallace V.},
  year={1978},
  publisher={Consulting Psychologists Press},
  address={Palo Alto, CA},
  doi={10.1037/t27734-000},
  url={https://doi.org/10.1037/t27734-000}
}

@phdthesis{baker-shenk1983microanalysis,
  title={A microanalysis of the nonmanual components of questions in American Sign Language},
  author={Baker-Shenk, Charlotte L},
  year={1983},
  school={University of California, Berkeley},
  url={https://escholarship.org/uc/item/7b03x0tz}
}

@book{neidle2000syntax,
  title={The Syntax of American Sign Language: Functional Categories and Hierarchical Structure},
  author={Neidle, Carol and Kegl, Judy and MacLaughlin, Dawn and Bahan, Benjamin and Lee, Robert G},
  year={2000},
  publisher={MIT Press},
  isbn={9780262140676}
}

@article{lin2019interrogative,
  author    = {Lin, Hao},
  title     = {Interrogative marking in {Chinese} {Sign} {Language}: {A} preliminary corpus-based investigation},
  journal   = {Sign Language \& Linguistics},
  year      = {2019},
  volume    = {22},
  pages     = {241--266},
  publisher = {John Benjamins Publishing Company},
  doi       = {10.1075/sll.19001.lin},
  url       = {https://doi.org/10.1075/sll.19001.lin}
}

@article{benitez2016the,
title = {The not face: A grammaticalization of facial expressions of emotion},
journal = {Cognition},
volume = {150},
pages = {77--84},
year = {2016},
issn = {0010-0277},
doi = {10.1016/j.cognition.2016.02.004},
url = {https://www.sciencedirect.com/science/article/pii/S0010027716300324},
author = {C. Fabian Benitez-Quiroz and Ronnie B. Wilbur and Aleix M. Martinez},
}

@misc{wfd2026faqs,
  author = {{World Federation of the Deaf}},
  title = {Frequently Asked Questions},
  year = {2026},
  url = {https://wfdeaf.org/contact/faqs/},
  note = {Accessed: 2026-08-27}
}

@InProceedings{perez2018film,
  title={FiLM: Visual Reasoning with a General Conditioning Layer},
  author={Ethan Perez and Florian Strub and Harm de Vries and Vincent Dumoulin and Aaron C. Courville},
  booktitle={Proceedings of the AAAI Conference on Artificial Intelligence},
  year={2018},
  doi={10.1609/aaai.v32i1.11671},
  url={https://aaai.org/papers/11671-film-visual-reasoning-with-a-general-conditioning-layer/}
}

\newpage
\appendix

\section*{Appendix}
\label{sec:appendix}

This supplementary material is organized as follows. Appendix~\ref{sec:domain} validates the FER-to-Sign domain transfer; Appendix~\ref{sec:implementation} provides extended implementation details; and Appendix~\ref{sec:extra-experiments} presents additional experiments and qualitative results.

\section{Domain Transfer Validation}
\label{sec:domain}

\subsection{Kinematic Overlap Between FER and Grammatical NMS}
\label{sec:fer2sign}

\paragraph{Bridging Affective and Grammatical Facial Dynamics.}
The facial-expression encoder, fine-tuned on FER datasets, serves as a domain-transferred feature extractor that learns an expression-sensitive latent manifold.
This manifold preserves fine-grained variations in facial configuration that may be informative for syntactic distinctions in SL.
The FER pretraining objective compels the encoder to discriminate subtle facial appearance changes across emotional states, which necessitates high sensitivity to the muscle activation patterns that constitute facial expressions. 
These patterns are canonically catalogued as Facial Action Units (AUs) by FACS~\cite{ekman1978facs}, and they constitute shared kinematic primitives between affective expressions and grammatical NMS in SL~\cite{baker-shenk1983microanalysis, neidle2000syntax, pfau2010nonmanuals}. 
Thus, AU annotations serve as a conceptual bridge to articulate the physical substrate common to both domains. 
The facial-dynamics prior acquired through FER pretraining may therefore facilitate transfer to expression variations that serve grammatical functions in SL.

Tables~\ref{tab:fer_au} and~\ref{tab:asl_nmm} theoretically delineate these shared kinematic primitives, mapping canonical FACS AU combinations for representative FER categories to the AUs associated with grammatical NMS in ASL and related Western sign languages~\cite{baker-shenk1983microanalysis, neidle2000syntax, pfau2010nonmanuals}. 
The mappings identify partial AU overlaps without assuming a one-to-one correspondence between emotion categories and grammatical functions: AU1+2 (brow raise) appears in both the surprise prototype and yes/no interrogatives; AU4 (brow furrow) is shared among anger, fear, wh-questions, and epistemic marking; and AU9 (nose wrinkler) participates in both disgust and negation~\cite{benitez2016the}.
Because grammatical markers often recruit localized movements while emotional expressions coordinate multiple facial regions, the transfer operates over continuous facial dynamics and does not use AU supervision or categorical emotion-to-grammar mappings.

\begin{table}[t]
\centering
\resizebox{\linewidth}{!}{%
\begin{tabular}{ll}
\toprule
\textbf{FER Category} & \textbf{Canonical AU Combination} \\
\midrule
Surprise & AU1+AU2, AU5, AU26 \\
Anger & AU4, AU5, AU7, AU23 \\
Happiness & AU6, AU12 \\
Sadness & AU1, AU4, AU15 \\
Fear & AU1+AU2, AU4, AU5, AU7, AU20, AU26 \\
Disgust & AU9, AU15, AU16 \\
Neutral & -- \\
\bottomrule
\end{tabular}%
}
\caption{Canonical FACS AU combinations for representative FER categories, illustrating the facial muscle patterns underlying emotional expressions.}
\label{tab:fer_au}
\end{table}

\begin{table}[t]
\centering
\resizebox{\linewidth}{!}{%
\begin{tabular}{ll}
\toprule
\textbf{Grammatical Function} & \textbf{Canonical AU Combination} \\
\midrule
Yes/no interrogatives & AU1+2, AU5 \\
Wh-questions \& emphasis & AU4 \\
Negative polarity & AU4, AU5, AU7, AU9 \\
Conditional & AU1+2 \\
Epistemic & AU4 \\
Strong negation & AU9, AU10 \\
Declarative baseline & -- \\
\bottomrule
\end{tabular}%
}
\caption{Canonical grammatical NMS in ASL and their associated AU combinations, describing the shared muscle primitives that bridge affective and grammatical facial dynamics.}
\label{tab:asl_nmm}
\end{table}

\begin{figure}[!t]
\centering
\includegraphics[width=\linewidth]{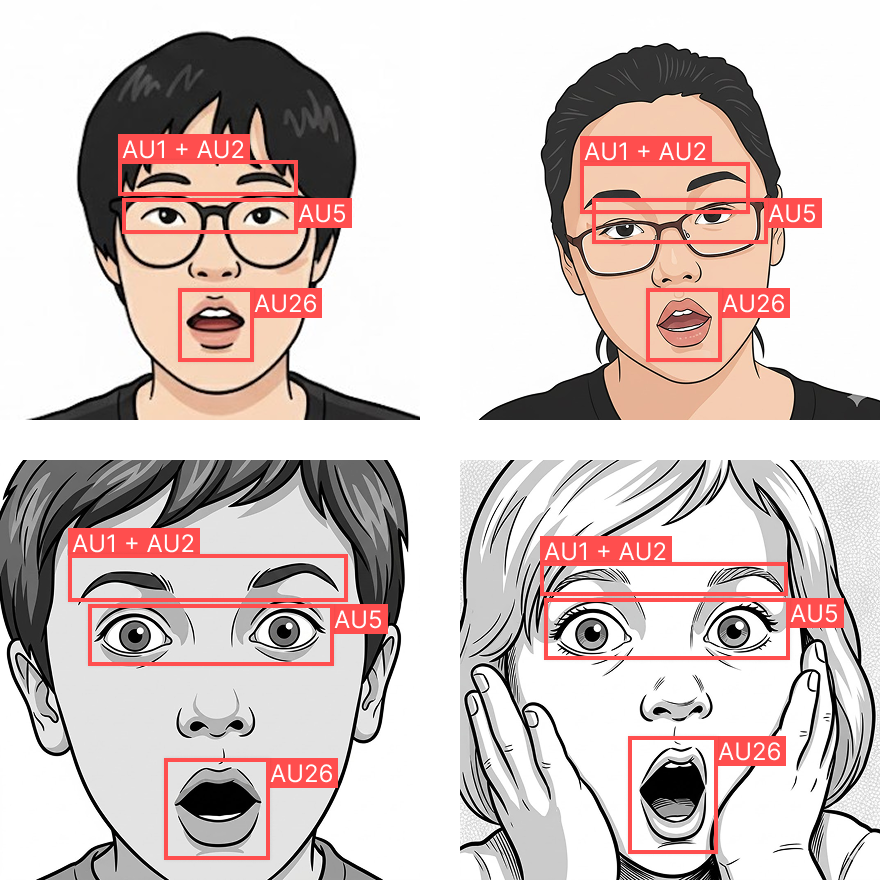}
\caption{Partial AU-level overlap between interrogative facial expressions in CSL‑Daily (top) and surprise prototypes in FER2013 (bottom). Shared patterns such as brow raise, upper-lid raise, and jaw drop motivate expression-sensitive cross-domain transfer.}
\label{fig:au}
\end{figure}

\paragraph{Case Study: Interrogative Constructions in CSL.}
To examine the plausibility of this kinematic bridge, we consider interrogative constructions in CSL as a concrete case study.
Corpus-based research by~\citet{lin2019interrogative} identifies brow raise (AU1+AU2) as a primary prosodic marker in CSL interrogatives. 
Figure~\ref{fig:au} illustrates the partial overlap between facial configurations in these interrogative contexts and the FER2013 \textit{surprise} prototype, including brow raise, upper-lid raise, and jaw drop.
This partial kinematic overlap motivates transfer of an expression-sensitive prior without requiring sign-specific facial labels.

\subsection{Facial Representation Validation}
\label{subsec:feature_validation}

To examine the structure of FER-derived facial representations across sentence categories, we project frame-level facial embeddings from a stratified CSL-Daily subset into a two-dimensional space using UMAP~\cite{mcinnes2018umap}, configured with $n\_neighbors=15$ and $min\_dist=0.1$ to balance local temporal coherence and global structural preservation.

Since the original dataset lacks explicit syntactic annotations, we construct sentence categories via a heuristic pipeline: (i) extracting punctuation patterns and interrogative lexical cues from spoken translations; (ii) applying rule-based filtering to identify candidate interrogative, negative, declarative, and mixed utterances; and (iii) conducting manual verification to resolve ambiguities. This process yields 400 validated samples evenly distributed across the four categories, where \textit{mixed} denotes utterances containing an interrogative clause interleaved with a declarative or negative clause (see Appendix~\ref{subsec:quantitative} for details). Representative examples are presented in Table~\ref{tab:syntactic-samples}, with English translations provided for reference.

\begin{table}[!t]
\centering
\footnotesize
\renewcommand{\arraystretch}{0.5}
\setlength{\tabcolsep}{4pt}
\begin{tabular}{@{} p{0.95\columnwidth} @{}}
\toprule
\multicolumn{1}{c}{\cellcolor{SkyBlue!50}\textbf{Negative}} \\
\midrule
他的声音不亲切。 \\
(His voice is not friendly.) \\
\midrule
不要在下方发弹幕。 \\
(Do not post comments below.) \\
\midrule
每天喝饮料对身体不好。 \\
(Drinking beverages daily is unhealthy.) \\
\midrule
\multicolumn{1}{c}{\cellcolor{green!20}\textbf{Declarative}} \\
\midrule
超市为人们的生活提供了便利。 \\
(Supermarkets provide convenience for people's daily lives.) \\
\midrule
他开了一家店出租礼服，生意十分兴隆。 \\
(He opened a shop renting formal wear, and business is thriving.) \\
\midrule
他努力使自己保持清醒的头脑。 \\
(He strives to keep a clear mind.) \\
\midrule
\multicolumn{1}{c}{\cellcolor{red!20}\textbf{Interrogative}} \\
\midrule
你看见小张了吗？ \\
(Have you seen Xiao Zhang?) \\
\midrule
我的杯子在哪里？ \\
(Where is my cup?) \\
\midrule
中国菜好吃吗？ \\
(Is Chinese food delicious?) \\
\midrule
\multicolumn{1}{c}{\cellcolor{blue!20}\textbf{Mixed}} \\
\midrule
不辣的菜，有没有？推荐一下。 \\
(Non-spicy dishes, do you have any? Please recommend.) \\
\midrule
我想请你看电影，你什么时候有时间？ \\
(I'd like to invite you to a movie, when are you free?) \\
\midrule
这条裤子很好看，你觉得怎么样？ \\
(These pants look nice, what do you think?) \\
\bottomrule
\end{tabular}
\caption{Representative utterances from the syntactically stratified CSL-Daily subset.}
\label{tab:syntactic-samples}
\end{table}

\begin{figure}[t]
    \centering
    \includegraphics[width=\linewidth]{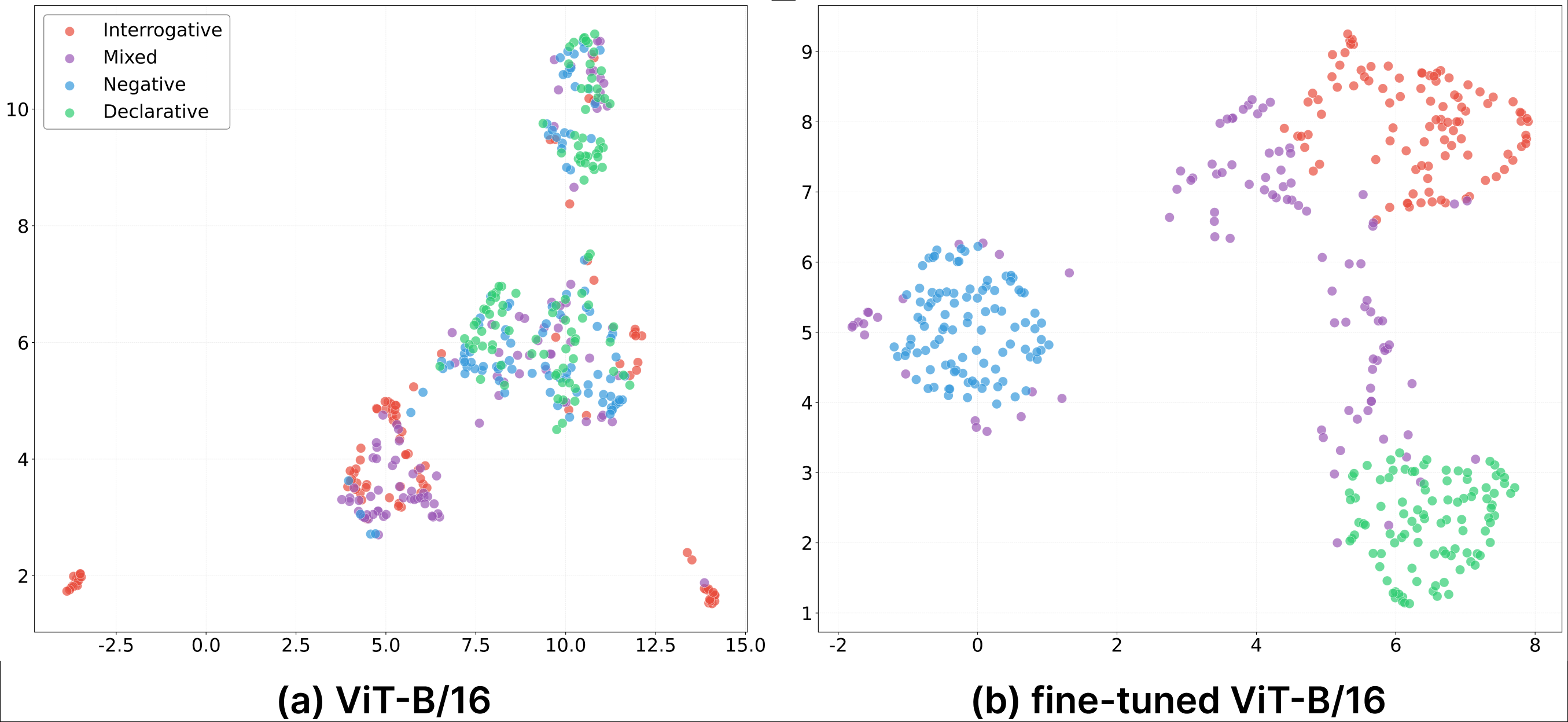}
    \caption{UMAP visualization of facial-expression embeddings from a syntactically stratified CSL-Daily subset. (a)~An off-the-shelf ViT-B/16 pretrained on ImageNet yields partial clustering with noticeable overlap among categories. (b)~The same architecture after fine-tuning on FER2013 produces tighter clusters for \colorbox{SkyBlue!50}{negative}, \colorbox{green!20}{declarative}, \colorbox{red!20}{interrogative}, and \colorbox{blue!20}{mixed} utterances. This separation provides supporting evidence that FER adaptation retains facial variation associated with these categories.}
    \label{fig:umap}
\end{figure}

As illustrated in Figure~\ref{fig:umap}, we compare the geometric structure of embeddings from two ViT-B/16 extractors. 
Panel~(a) uses an off-the-shelf ViT-B/16 pretrained on ImageNet~\cite{dosovitskiy2021image}: while the four syntactic categories exhibit coarse spatial grouping, substantial inter-category overlap persists.
In contrast, panel~(b) employs the same architecture fine-tuned on FER2013~\cite{trpakov2023vitface}. Here, clusters become notably tighter and more separable, with clearer geometric boundaries across categories. 
Mixed utterances occupy intermediate regions between interrogative and declarative or negative clusters, and residual overlap remains across categories.
This structure emerges without syntactic supervision during facial encoder training and is consistent with FER priors retaining facial variation associated with these sentence categories.
These observations support the utility of FER-derived facial dynamics as complementary cues in gloss-free SLT.

\section{More Implementation Details}
\label{sec:implementation}

\subsection{Temporal Modeling}
\label{sec:temporal_modeling}
For short-term modeling of multimodal sequences, we employ a 1D TCN~\cite{bai2018tcn} with the architecture \{K5, P2, K5, P2\}, where K$\sigma$ denotes a kernel size of $\sigma$ and P$\sigma$ indicates a pooling layer with kernel size $\sigma$ \cite{hu2023continuous}. This configuration captures local motion patterns while reducing sequence length. The features obtained after temporal modeling are integrated into the LLM's embedding space via a cross-modal MLP connector \cite{liu2024improved} with two hidden layers.

\subsection{Prompt Design}
\label{sec:prompt_design}
Following prior SLT works, we employ in-context learning~\cite{brown2020language} with a structured multilingual prompt template. For each training example, we first translate the text into multiple languages (e.g., English, French, and Spanish) using professional translation services. Table \ref{tab:prompt_template} shows our prompt template design. During training, we randomly shuffle the in-context examples within each batch to ensure contextual independence from the target translation. This prevents the model from memorizing specific example-target mappings. At inference time, we hard-code a fixed set of in-context examples sampled from the training set to maintain consistency across evaluations and ensure no data leakage.

\begin{table}[htbp]
\centering
\small
\begin{tabularx}{\linewidth}{>{\raggedright\arraybackslash}X}
\toprule
\texttt{[SIGN\_FEATURES]} Translate the given sentence into German. 
It can occasionally thunderstorms.=vereinzelt kann es gewittern.
Ocasionalmente puede tormentas eléctricas.=vereinzelt kann es gewittern.
Il peut parfois les orages.=vereinzelt kann es gewittern. \\
\midrule
\texttt{[SIGN\_FEATURES]} Translate the given sentence into Chinese. 
He left after eating his fill. = 他吃饱饭，就离开了。
Después de comer hasta saciarse, se fue. = 他吃饱饭，就离开了。
Après avoir mangé à sa faim, il est parti. = 他吃饱饭，就离开了。\\
\bottomrule
\end{tabularx}
\caption{Exemplary prompt template design for the PHOENIX14T (top) and CSL-Daily (bottom) datasets.}
\label{tab:prompt_template}
\end{table}

\subsection{More Dataset Details}
\label{sec:more_dataset_details}

\begin{figure}[htbp]
\centering
\includegraphics[width=\columnwidth]{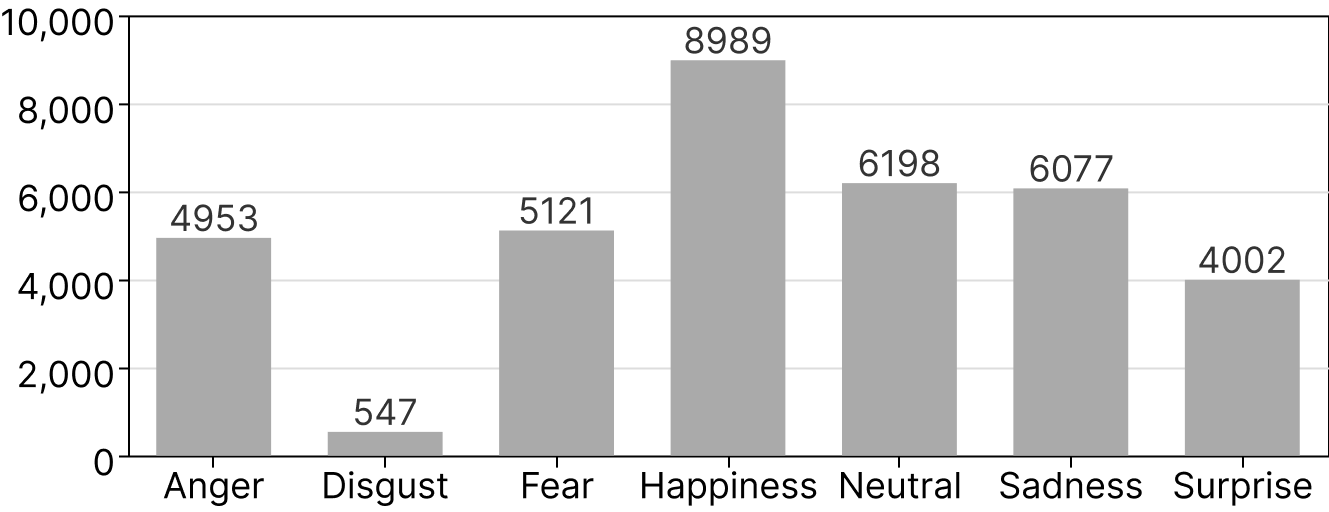}
\caption{Class distribution of the FER2013 facial expression dataset across seven categories (anger, disgust, fear, happiness, neutral, sadness, surprise).}
\label{fig:fer2013}
\end{figure}

\paragraph{FER Dataset.}
FER2013~\cite{goodfellow2013challenges} is a widely-used facial expression dataset containing 35,887 grayscale images of size 48$\times$48 pixels, divided into 28,709 training samples and 7,178 test samples. The dataset comprises seven basic facial expression categories: anger, disgust, fear, happiness, neutral, sadness, and surprise. Images in FER2013 were automatically collected from the web and labeled through crowdsourcing, making it a challenging benchmark due to variations in lighting conditions, head poses, and partial occlusions. Figure~\ref{fig:fer2013} illustrates the class distribution of images in FER2013 across the seven facial expression categories.

\paragraph{Usage License.}
We comply with the licensing terms of all utilized datasets. PHOENIX14T~\cite{camgoz2018neural} is distributed under CC BY-NC-SA 3.0, FER2013~\cite{goodfellow2013challenges} under CC BY 4.0, and CSL-Daily~\cite{zhou2021improving} is used with explicit written permission from the authors in accordance with their stated usage agreements.

\subsection{Preprocessing Pipeline}
\label{sec:preprocess_details}
To ensure high-fidelity feature representation and reproducibility, all sign language videos are processed through a standardized pipeline using frozen, off-the-shelf backbones. We utilize the HuggingFace Hub as the primary model repository, employing each processor's native normalization and resizing protocols. All our usage complies with the requirements of the open-source licenses of the respective models.
Spatial features employ CLIP-ViT-L/14~\cite{radford2021learning}\footnote{\url{https://huggingface.co/openai/clip-vit-large-patch14}}, while motion features utilize VideoMAE-L/16~\cite{tong2022videomae}\footnote{\url{https://huggingface.co/MCG-NJU/videomae-large}}. 
Facial-expression features (FE) are extracted using a ViT-B/16~\cite{dosovitskiy2021image, trpakov2023vitface} fine-tuned on FER2013\footnote{\url{https://huggingface.co/trpakov/vit-face-expression}}. 
We strictly adhere to the official split protocols for each dataset. For temporal consistency, PHOENIX14T \cite{camgoz2018neural} videos are processed at their native 25 FPS, while CSL-Daily \cite{zhou2021improving} videos are sampled at 30 FPS.

\subsection{Evaluation Metrics}
\label{sec:metrics_details}

Our evaluation framework employs three standard metrics for SLT assessment: BLEU-n \cite{papineni2002bleu}, ROUGE-L \cite{lin2004rouge}, and BLEURT \cite{sellam2020bleurt}. 
Prior to the evaluation, adhering to community standards, we apply text normalization (lowercasing and punctuation removal) to PHOENIX14T. For CSL-Daily, we perform character-level evaluation while preserving original punctuation to accurately reflect Chinese linguistic structures. 
Below we detail their mathematical formulations and implementation details.

\textbf{BLEU-n} measures translation quality by calculating the precision of n-gram matches between predictions and references. For each order $n$ (typically 1-4), the score is computed as:
\begin{equation}
\text{BLEU-n} = BP \cdot \exp\left(\frac{1}{N}\sum_{n=1}^{N}\log p_n\right)
\end{equation}
where $p_n$ is the modified n-gram precision, $BP$ is the brevity penalty that penalizes overly short translations, and $N$ is the maximum n-gram order. In our implementation, we use the \texttt{sacrebleu} \cite{post2018call} library's BLEU metric with language-specific tokenization: character-level tokenization for Chinese (using the \texttt{zh} tokenizer) and the standard \texttt{13a} tokenizer for German text. This adaptation ensures appropriate handling of morphological differences across languages.

\textbf{ROUGE-L} evaluates translation quality based on the longest common subsequence (LCS) between the predicted sequence $X$ and reference sequence $Y$. The metric computes precision, recall, and their harmonic mean (F1 score) as follows:
\begin{align}
\text{Precision} &= \frac{\operatorname{LCS}(X,Y)}{|X|}, \\
\text{Recall} &= \frac{\operatorname{LCS}(X,Y)}{|Y|}, \\
\text{ROUGE-L} = \text{F1} &= \frac{2 \cdot \text{Precision} \cdot \text{Recall}}{\text{Precision} + \text{Recall}},
\end{align}
where $|\cdot|$ denotes sequence length. Scores are computed using the \texttt{rouge\_score} library.

\textbf{BLEURT} extends traditional metrics by leveraging contextual embeddings from BERT \cite{devlin2019bert} to assess semantic equivalence. We implement BLEURT using the \texttt{BLEURT-20} checkpoint~\cite{sellam2020bleurt}\footnote{\url{https://github.com/google-research/bleurt}}, which was fine-tuned on human judgments and demonstrates strong correlation with human evaluations across multiple languages. 
Specifically, BLEURT-20 has been tested on 13 languages: Chinese, Czech, English, French, German, Japanese, Korean, Polish, Portuguese, Russian, Spanish, Tamil, and Vietnamese~\cite{pu2021learning}.
The metric computes a regression score based on the embedding similarity between predictions and references, ranging from 0 to 1, thereby capturing semantic nuances that may be overlooked by n-gram-based metrics. 
Following prior work, we report BLEURT scores scaled by $\times 100$ for readability.

\subsection{Hyperparameters and Resources}
\label{sec:hyperparameters}

\paragraph{Hyperparameters.}
We report the hyperparameter configurations used in all experiments. For motion feature extraction, we adopt a sliding window with width $w=16$ and stride $s_m=8$, yielding a motion feature sequence of length $T_m = \lfloor (T - w) / s_m \rfloor + 1$ for a video of $T$ frames. FE are temporally downsampled with interval $s_e=8$ to mitigate redundancy, resulting in a sequence of length $T_e = \lfloor (T - 1) / s_e \rfloor + 1$. All features from frozen encoders are projected through a lightweight head with hidden dimension $d = 768$.

The FEAM module employs bidirectional Modulator submodules, each implemented as a two-layer MLP with GELU activation and dropout rate $0.1$. The parameter predictor processes concatenated inputs $[\mathbf{Z}_q, \mathbf{Z}_c^{\text{a}}]$ with hidden dimension $2d$; we initialize the final linear layer to zero to stabilize early optimization. The gating network projects to $d/2$ before outputting a scalar gate via Sigmoid. Learnable modality embeddings for spatial, spatiotemporal, and facial streams are drawn from $\mathcal{N}(0, 0.02)$. The fusion weight $w_{\text{p}}$ for aggregating reverse-modulated facial features is initialized to zero, and the modulation magnitude $\alpha$ is fixed at $0.5$.

For LLM fine-tuning, we apply LoRA with rank $16$, scaling factor $32$, and dropout $0.1$. Optimization uses AdamW~\cite{loshchilov2019decoupled} with $(\beta_1, \beta_2) = (0.9, 0.98)$ and weight decay $0.01$. We adopt a cosine learning rate schedule with linear warmup over the first 10\% of steps, peaking at $6 \times 10^{-4}$. Label smoothing~\cite{szegedy2016rethinking} with $\epsilon = 0.1$ is applied to output logits to mitigate overfitting. All experiments fix the random seed to 0 for reproducibility.

On PHOENIX14T, we train for 500 epochs with batch size 8 (gradient accumulation steps $= 2$) and use beam search of width 5 during inference. For CSL-Daily, we set batch size to 4, peak learning rate to $1 \times 10^{-4}$, and train for 200 epochs. The contrastive loss weight $\lambda$ is set to 1.0 for both datasets.

\paragraph{Resources.}
All experiments are conducted on a single NVIDIA A100 (80GB) GPU. Our implementation is based on PyTorch 2.0~\cite{pytorch2019paszke} with CUDA 12.8 and employs bf16 mixed-precision training.

Regarding computational efficiency, our model contains 3.0B total parameters, of which only 52.4M (roughly 1.7\%) are trainable. The model converges efficiently, typically reaching near-optimal performance within 4K and 8K training steps without needing to reach the maximum number of epochs, corresponding to approximately 12 and 24 hours on PHOENIX14T and CSL-Daily, respectively.

\subsection{Handling Detection Failures}
\label{sec:failure}

We employ RetinaFace~\cite{deng2019retinaface} for high-precision face localization, achieving 100\% video-level detection across all dataset splits. Frame-level misses occur at negligible rates on both datasets: 0.44\%, 0.45\%, and 0.41\% for the training, development, and test sets of PHOENIX14T, respectively, and 0.04\%, 0.03\%, and 0.04\% for the corresponding splits of CSL-Daily. The significantly lower failure rate on CSL-Daily can be attributed to its controlled recording environment. Manual inspection confirms that these occasional failures stem primarily from self-occlusion and extreme head poses during dynamic signing.
To preserve temporal continuity without introducing distributional shifts from non-facial regions, we reconstruct missing facial features via distance-weighted interpolation. For an undetected frame $t$, the smoothed representation $\hat{\mathbf{z}}_t^e$ is computed from its nearest successfully detected neighbors $\mathcal{N}(t)$:
\begin{equation}
    \hat{\mathbf{z}}_e^t = \frac{\sum_{k \in \mathcal{N}(t)} \alpha_k \hat{\mathbf{z}}_e^k}{\sum_{k \in \mathcal{N}(t)} \alpha_k}, \alpha_k = \frac{1}{|t - k| + \epsilon},
\end{equation}
where $\epsilon=10^{-6}$ ensures numerical stability. This interpolation strategy effectively bridges transient detection gaps while maintaining the temporal coherence of non-manual cues.

\begin{figure}[htbp]
\centering
\includegraphics[width=\columnwidth]{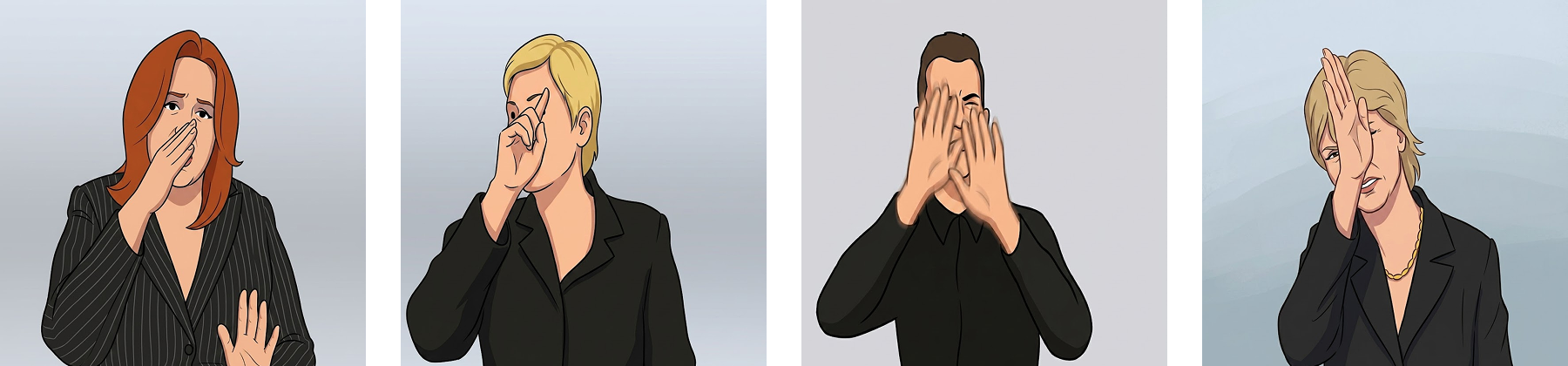}
\caption{Representative face detection failure cases on PHOENIX14T, illustrating challenges from self-occlusion and extreme head poses.}
\label{fig:failure}
\end{figure}

\section{More Experiments}
\label{sec:extra-experiments}

\subsection{Ablation Study for CSL-Daily}
\label{subsec:csl_ablation}
To evaluate the cross-dataset generalizability of the proposed modules and assess the effectiveness of facial-expression modeling in CSL scenarios, we conduct ablation experiments on the CSL-Daily test set. We examine two configurations: removal of the FE feature branch and exclusion of the FEAM module. The configuration without FE features is a within-framework SpaMo-style spatial-motion baseline under the same controlled configuration, enabling an isolated comparison that quantifies the performance gain attributable to explicit facial-expression modeling. Results are presented in Table~\ref{tab:csl_ablation}.

Adding FE through direct concatenation improves B-4 from 21.48 to 21.94 and R-L from 48.66 to 49.56, showing measurable gains from facial features without FEAM. Adding FEAM further raises B-4 to 22.94 and R-L to 50.40, indicating an additional benefit from bidirectional fusion. This incremental comparison separates the contribution of facial features from that of FEAM.

These findings mirror the ablation results on PHOENIX14T and show consistent trends across the two evaluated sign languages. The BLEU‑4 score improvement on CSL‑Daily (6.80\%) is slightly higher than that on PHOENIX14T (6.54\%).

\begin{table}[htbp]
\centering
\footnotesize
\renewcommand{\arraystretch}{0.8}
\setcounter{ablationrow}{0}
\resizebox{\linewidth}{!}{
    \begin{tabular}{ccccccccc}
    \toprule
     & \textbf{FE} & \textbf{FEAM} & B-1 & B-2 & B-3 & B-4 & R-L \\
    \midrule
    \ablnum & -- & -- & 48.67 & 35.97 & 27.42 & 21.48 & 48.66 \\
    \ablnum & \checkmark & -- & 49.01 & 36.36 & 27.88 & 21.94 & 49.56 \\
    \dashedmidrule
    \ablnum & \checkmark & \checkmark & \textbf{50.83} & \textbf{37.82} & \textbf{29.03} & \textbf{22.94} & \textbf{50.40} \\
    \bottomrule
    \end{tabular}
}
\caption{Ablation study of core components on the CSL-Daily test set. \checkmark~indicates the component is included.}
\label{tab:csl_ablation}
\end{table}

\subsection{Statistical Significance Analysis}
\label{sec:significance}
To rigorously assess the contribution of facial-expression modeling, we conduct statistical significance testing between our full model and an ablated variant excluding FE. We employ 10{,}000 bootstrap resampling iterations on fixed-model test-set translations to compute two-tailed p-values and 95\% confidence intervals for BLEU-4 score differences.

\begin{figure}[htbp]
\centering
\includegraphics[width=\linewidth]{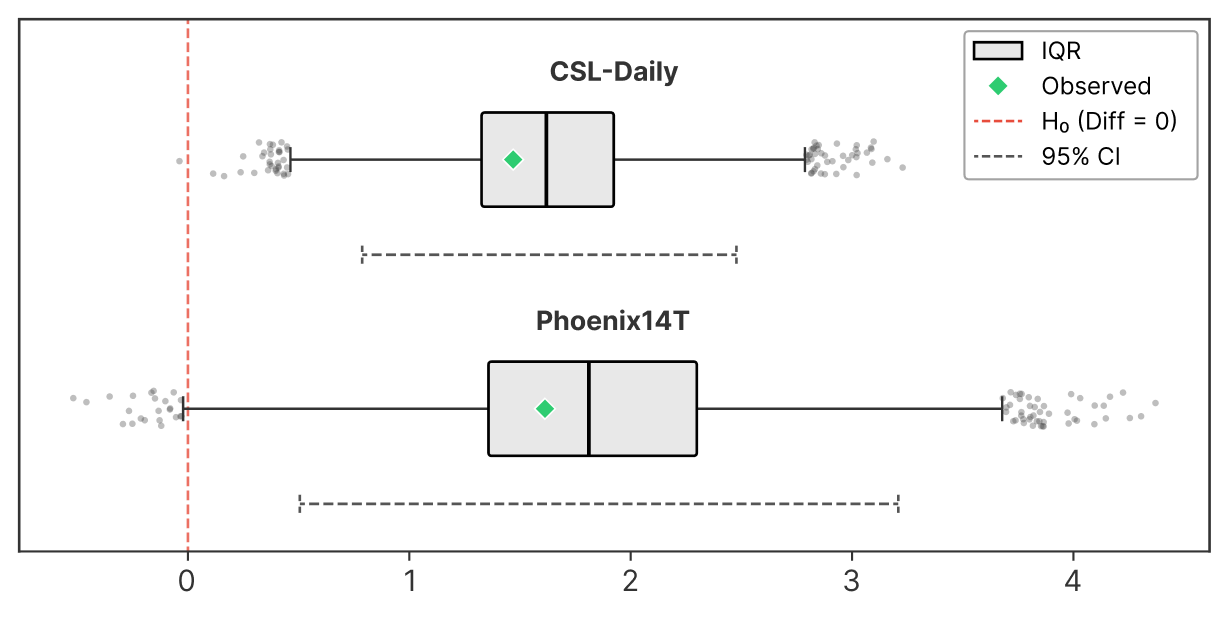}
\caption{Statistical significance analysis via bootstrap resampling of fixed-model outputs (10,000 iterations). Box plots show BLEU-4 score differences between FEA-SLT and its ablated variant (w/o FE) on CSL-Daily and PHOENIX14T. Green diamonds indicate observed performance gains. Dashed lines mark 95\% confidence intervals. The output-level improvements are statistically significant (p<0.01).}
\label{fig:significance}
\end{figure}

The resulting distributions are visualized as box plots in Figure~\ref{fig:significance}, where the boxes delineate the interquartile range (IQR) with median lines, green diamonds indicate the observed performance gains, and dashed lines represent the 95\% confidence intervals. Results indicate statistically significant improvements across both evaluation datasets. On CSL-Daily, incorporating FE and FEAM yields a +1.46 BLEU-4 improvement with strong significance (p=0.0002, 95\% CI [0.79, 2.48]). Similarly, PHOENIX14T exhibits a +1.62 BLEU-4 gain that reaches statistical significance (p=0.0054, 95\% CI [0.51, 3.21]). In both cases, the confidence intervals lie entirely above zero, providing output-level evidence of improvements on the two evaluated test sets.

\subsection{Facial-Sensitive Subset Evaluation}
\label{subsec:ambiguity_subset}

CSL-Daily does not provide controlled annotations for pairs with matched MS and contrasting facial expressions. We therefore curate a targeted facial-sensitive subset from the test set to further evaluate facial-expression modeling.

The subset contains 84 instances selected through keyword filtering for affective predicates (e.g., ``害怕'' \textit{afraid}, ``生气'' \textit{angry}) and interrogative markers, which are frequently accompanied by distinctive facial-expression cues in SL. We then manually verify that the selected samples exhibit visually salient facial expressions associated with affective or interrogative semantics. Representative examples are shown in Table~\ref{tab:ambiguity-examples}. We use this manually verified subset as supporting diagnostic evidence for facial-sensitive cases. Notably, the performance gains on this subset are substantially larger than those observed on the full CSL-Daily test set, suggesting that facial-expression modeling is particularly useful when facial cues are semantically or syntactically salient.

\begin{figure}[!h]
\centering
\includegraphics[width=\columnwidth]{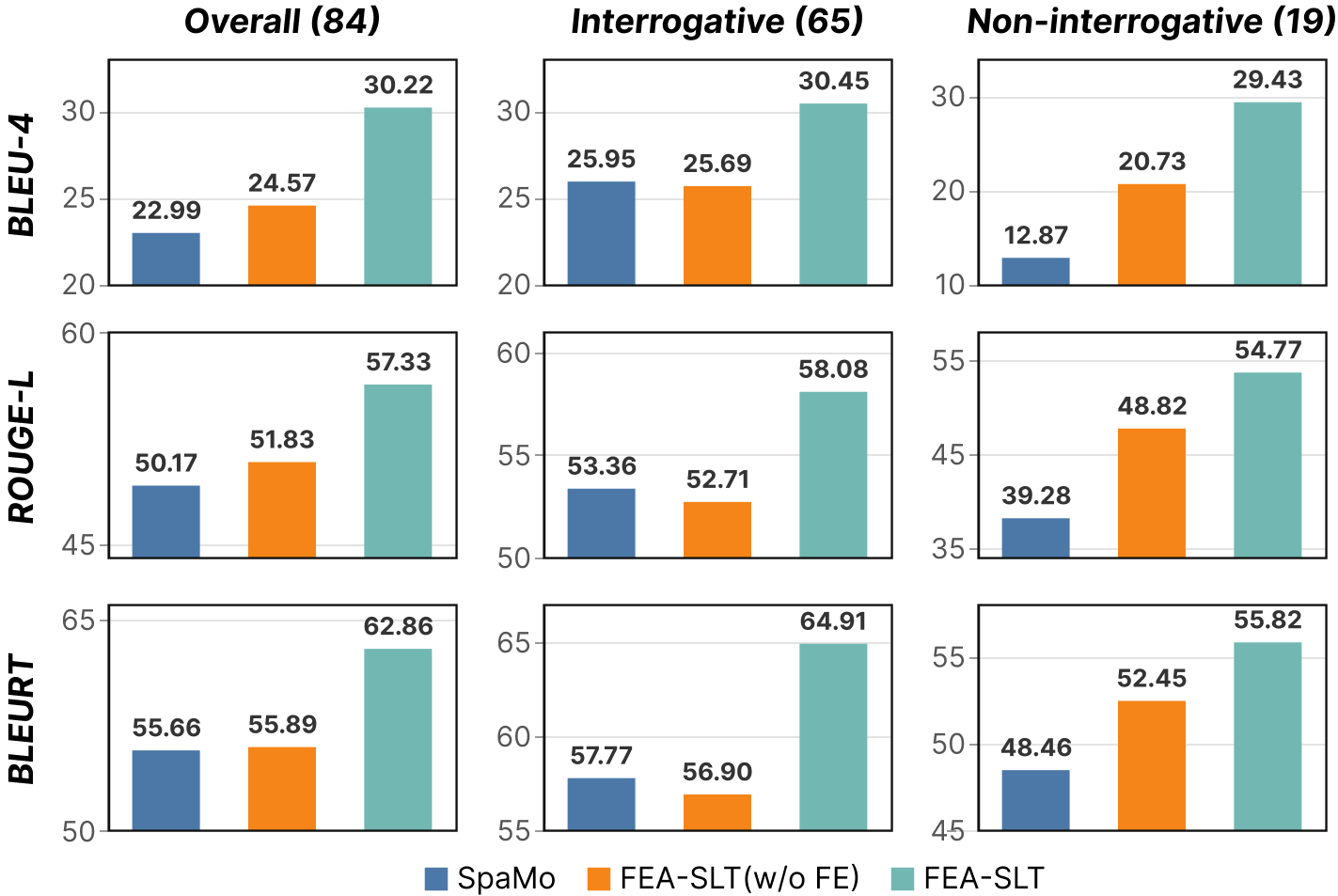}
\caption{Controlled model comparison on the curated 84-instance facial-sensitive subset and its interrogative (65) and non-interrogative (19) partitions. Rows report BLEU-4, ROUGE-L, and BLEURT.}
\label{fig:subset}
\end{figure}

We perform a controlled model comparison on this fixed 84-instance subset using BLEU-4, ROUGE-L, and BLEURT. Specifically, we evaluate SpaMo, FEA-SLT without the FE branch, and the full FEA-SLT model on the same examples. As shown in Figure~\ref{fig:subset}, FEA-SLT consistently outperforms both baselines across all metrics. In particular, BLEU-4 improves from 22.99 (SpaMo) and 24.57 (w/o FE) to 30.22. ROUGE-L improves from 50.17 (SpaMo) and 51.83 (w/o FE) to 57.33, while BLEURT improves from 55.66 (SpaMo) and 55.89 (w/o FE) to 62.86. Compared with FEA-SLT without FE, the full model gains +5.65 BLEU-4, +5.50 ROUGE-L, and +6.97 BLEURT, isolating the contribution of explicit facial-expression modeling within the shared backbone. These results support the claim that facial-expression features provide complementary facial-sensitive cues beyond manual signing alone, particularly for instances involving affective or interrogative semantics.

We further split the subset into 65 interrogative and 19 non-interrogative facial-sensitive samples (Figure~\ref{fig:subset}); the latter include manually checked affective cases such as ``afraid,'' ``angry,'' and ``sad.'' Relative to FEA-SLT without FE, the full model improves BLEURT/BLEU-4/ROUGE-L by +8.01/+4.76/+5.37 on interrogative samples and +3.37/+8.70/+5.95 on non-interrogative samples. BLEURT is reported on a \(0\)--\(100\) scale throughout. The gains in both groups support benefits across interrogative and non-interrogative facial-sensitive cases.

\begin{table}[h]
\centering
\footnotesize
\renewcommand{\arraystretch}{1}
\setlength{\tabcolsep}{5pt}
\begin{tabular}{p{0.9\linewidth}}
\toprule
天黑了，我\cjkbold{害怕}。 \\
(It is dark, I am \textbf{afraid}.) \\
\midrule
我喜欢听惊险的故事，但有时越听越\cjkbold{害怕}。 \\
(I enjoy thrilling stories, but sometimes they make me feel more and more \textbf{afraid}.) \\
\midrule
有些小孩子很\cjkbold{害怕}输液，一直\cjkbold{嚎啕大哭}。 \\
(Some children are \textbf{afraid} of infusions and keep \textbf{crying loudly}.) \\
\midrule
我很难相处，所以别惹我\cjkbold{生气}。 \\
(I am hard to get along with, so do not make me \textbf{angry}.) \\
\midrule
那些遇难者的家属，每个人都\cjkbold{流泪}\cjkbold{难过}。 \\
(Family members of the victims are all \textbf{shedding tears} and feeling \textbf{sorrowful}.) \\
\midrule
中午去\textbf{哪里}吃饭，在学校还是去饭店\textbf{?} \\
(\textbf{Where} shall we have lunch, at school or at a restaurant\textbf{?}) \\
\midrule
椅子上有一件衣服，是\textbf{谁}的\textbf{?} \\
(There is a piece of clothing on the chair; \textbf{whose} is it\textbf{?}) \\
\bottomrule
\end{tabular}
\caption{Representative examples from the facial-sensitive subset. Boldface highlights affective predicates or interrogative markers that are frequently associated with facial-expression cues in sign languages.}
\label{tab:ambiguity-examples}
\end{table}

\subsection{Ablation Study on Spatial Multi-Scale ($\text{S}^2$) Scaling}
\label{subsec:s2_scaling_ablation}

In the spatial feature extraction stage, we adopt the $\text{S}^2$ multi-scale strategy \cite{shi2024when}, which mitigates information loss in localized regions by integrating global contextual cues with local high-resolution details. To validate this design, we conduct supplementary ablation experiments on the PHOENIX14T test set, comparing three configurations: utilizing solely global features, directly concatenating global features with all four local branch features, and our adopted $\text{S}^2$ strategy that aggregates local features via averaging before fusion. The quantitative results are summarized in Table~\ref{tab:s2_scaling_ablation}.

As shown in the table, the exclusive use of global features yields suboptimal performance, indicating that coarse representations struggle to capture the fine-grained manual configurations essential for sign semantics. Direct concatenation of local branches introduces substantial feature redundancy, which marginally degrades translation quality despite retaining spatial details. In contrast, the $\text{S}^2$ strategy effectively balances multi-scale representation by averaging local features prior to fusion. This approach preserves discriminative local cues while maintaining a compact feature space, achieving consistent improvements across all evaluation metrics.

\begin{table}[htbp]
\centering
\footnotesize
\renewcommand{\arraystretch}{0.8}
\resizebox{\linewidth}{!}{
    \begin{tabular}{lcccccc}
    \toprule
    \textbf{Configuration} & B-1 & B-2 & B-3 & B-4 & R-L \\
    \midrule
    Global only & 51.67 & 39.09 & 31.12 & 25.74 & 47.74 \\
    Global concat Local & 51.71 & 38.94 & 30.98 & 25.61 & 47.97 \\
    \dashedmidrule
    \textbf{$\text{S}^\textbf{2}$-Wrapper} & \textbf{52.59} & \textbf{39.72} & \textbf{31.67} & \textbf{26.38} & \textbf{48.63} \\
    \bottomrule
    \end{tabular}
}
\caption{Ablation study of the $\text{S}^2$ multi-scale spatial feature extraction strategy on the PHOENIX14T test set.}
\label{tab:s2_scaling_ablation}
\end{table}

\subsection{Ablation Study on Fusion Architectures}
\label{subsec:fusion_arch_ablation}
The FEAM mechanism serves as the core component of our fusion architecture, adaptively integrating FE with spatiotemporal manual sign features. Motivated by the linguistic principle of manual and non-manual co-articulation in SL, FEAM employs a bidirectional modulation strategy to explicitly model the interdependence between facial and manual channels. To assess the contribution of this design, we conduct a controlled ablation study on the PHOENIX14T test set, comparing FEAM with several representative fusion paradigms: early fusion via concatenation, cross-attention for dynamic feature interaction, FiLM-based conditional modulation, and an SE-style channel attention mechanism. The Concat variant is a Sign Adaptor-style direct-fusion baseline corresponding to Row (2) of Table~\ref{tab:ablation}. All variants share the same facial-expression and spatiotemporal feature branches and training protocol. Quantitative results are summarized in Table~\ref{tab:fusion_arch_ablation}.

FEAM consistently achieves the highest performance across all evaluation metrics. These gains support the effectiveness of bidirectional modulation for integrating facial and manual features relative to the evaluated alternatives. The comparison controls the feature branches and training protocol, but not the parameter count.

\begin{table}[htbp]
\centering
\footnotesize
\renewcommand{\arraystretch}{0.8}
\resizebox{\linewidth}{!}{
    \begin{tabular}{lcccccc}
    \toprule
    \textbf{Configuration} & B-1 & B-2 & B-3 & B-4 & R-L \\
    \midrule
    Concat & 50.86 & 38.36 & 30.56 & 25.30 & 47.33 \\
    Cross-attention & 51.46 & 38.77 & 30.87 & 25.59 & 47.92 \\
    FiLM & 50.92 & 38.62 & 30.90 & 25.81 & 48.32 \\
    SE-like & 50.83 & 38.46 & 30.62 & 25.39 & 47.81 \\
    \dashedmidrule
    \textbf{FEAM} & \textbf{52.59} & \textbf{39.72} & \textbf{31.67} & \textbf{26.38} & \textbf{48.63} \\
    \bottomrule
    \end{tabular}
}
\caption{Comparison of fusion architectures on the PHOENIX14T test set.}
\label{tab:fusion_arch_ablation}
\end{table}

\subsection{Prompt Context}
\label{subsec:prompt-ablation}
We investigate the impact of contextual prompts elaborated in Appendix~\ref{sec:prompt_design} by comparing FEA-SLT with and without translation example contexts. Table~\ref{tab:prompt_ablation} reports B-1 to B-4 and R-L scores on the PHOENIX14T test set. Removing contextual prompts causes performance degradation across all metrics, confirming that in-context learning prompts enhance semantic alignment in SLT.

\begin{table}[htbp]
\centering
\footnotesize
\renewcommand{\arraystretch}{0.8}
\resizebox{\linewidth}{!}{
    \begin{tabular}{lcccccc}
    \toprule
    \textbf{Configuration} & B-1 & B-2 & B-3 & B-4 & R-L \\
    \midrule
    w/o context & 51.98 & 39.24 & 31.27 & 25.88 & 48.02 \\
    \dashedmidrule
    \textbf{w context} & \textbf{52.59} & \textbf{39.72} & \textbf{31.67} & \textbf{26.38} & \textbf{48.63} \\
    \bottomrule
    \end{tabular}
}
\caption{Ablation study on contextual prompts.}
\label{tab:prompt_ablation}
\end{table}

\subsection{Label Smoothing}
\label{subsec:labelsmoothing-ablation}
Label smoothing is a regularization technique that mitigates overconfidence in model predictions by replacing hard targets with smoothed distributions \cite{szegedy2016rethinking}. We investigate its impact on SLT by comparing FEA-SLT with and without label smoothing. Following standard practice, we set the smoothing parameter $\epsilon = 0.1$ to balance between preserving label information and reducing model overfitting. Table~\ref{tab:smooth_ablation} reports B-1 to B-4 and R-L scores on the PHOENIX14T test set. The results demonstrate consistent improvements across all metrics with label smoothing, especially in the R-L score, indicating enhanced generalization capability. This improvement is especially valuable for SLT where visual ambiguities often lead to prediction uncertainty.

\begin{table}[htbp]
\centering
\footnotesize
\renewcommand{\arraystretch}{0.8}
\resizebox{\linewidth}{!}{
    \begin{tabular}{lcccccc}
    \toprule
    \textbf{Configuration} & B-1 & B-2 & B-3 & B-4 & R-L \\
    \midrule
    w/o smoothing & 51.78 & 38.86 & 30.86 & 25.54 & 47.76 \\
    \dashedmidrule
    \textbf{w/ smoothing} & \textbf{52.59} & \textbf{39.72} & \textbf{31.67} & \textbf{26.38} & \textbf{48.63} \\
    \bottomrule
    \end{tabular}
}
\caption{Ablation study on label smoothing.}
\label{tab:smooth_ablation}
\end{table}

\subsection{LLM Architecture Selection}
\label{subsec:llm}

The selection of the backbone language model substantially influences SLT performance. As shown in Table~\ref{tab:llm_selection}, instruction-tuned models consistently achieve higher translation quality than their standard pretrained counterparts. Flan-T5-XL (3.0B) strikes an effective balance between translation accuracy and computational cost, outperforming the larger mT5-XL (3.7B) despite its smaller parameter count.

We observed that optimization stability within the mT5 family varies with model scale. At a uniform learning rate of $6\times10^{-4}$, mT5-xl and mT5-large exhibit training divergence, evidenced by unstable contrastive loss and degraded validation metrics, while mT5-base remains stable under identical conditions. These observations indicate that larger variants may require more careful optimization scheduling. Reducing the learning rate to $1\times10^{-4}$, which proved stable on CSL-Daily, stabilizes mT5-XL training on PHOENIX14T. Nevertheless, even with this adjustment, mT5-XL achieves BLEU-4 of 23.13 and ROUGE-L of 43.74, values that remain below those of Flan-T5-XL. These results suggest that the performance disparity reflects architectural and pretraining differences in addition to optimization factors.

We hypothesize that Flan-T5-XL's instruction-tuning objective aligns more closely with the directive nature of SLT, particularly when combined with task-specific prompts and in-context translation exemplars. This alignment may enhance the model's capacity to interpret and decode the soft prompts produced by the FEAF module.

\begin{table}[htbp]
\centering
\footnotesize
\setlength{\tabcolsep}{4pt}
\resizebox{\columnwidth}{!}{
    \begin{tabular}{lcccccc}
    \toprule
    \textbf{Model} & \textbf{Params} & B-1 & B-2 & B-3 & B-4 & R-L \\
    \midrule
    \multicolumn{7}{c}{\cellcolor{gray!10}\textbf{mT5 \cite{xue2021mt5}}} \\
    \midrule
    mT5-xl & 3.7B & 41.19 & 28.05 & 20.96 & 16.70 & 36.52  \\
    mT5-large & 1.2B & 20.31 & 10.41 & 6.85 & 5.26 & 14.19  \\
    mT5-base & 0.58B & 39.89 & 26.84 & 19.15 & 12.23 & 32.68  \\
    \midrule
    \multicolumn{7}{c}{\cellcolor{gray!10}\textbf{Flan-T5 \cite{chung2022flant5}}} \\
    \midrule
    \textbf{Flan-T5-xl} & 3.0B & \textbf{52.59} & \textbf{39.72} & \textbf{31.67} & \textbf{26.38} & \textbf{48.63} \\
    Flan-T5-large & 0.78B & 50.34 & 37.35 & 29.42 & 24.21 & 45.86  \\
    Flan-T5-base & 0.25B & 50.17 & 36.94 & 28.90 & 23.65 & 45.84  \\
    \midrule
    \multicolumn{7}{c}{\cellcolor{gray!10}\textbf{mBART \cite{liu2020multilingual,tang2020multilingual}}} \\
    \midrule
    mBART-large-50 & 0.6B & 47.74 & 35.00 & 27.27 & 22.17 & 43.87  \\
    mBART-large-cc25 & 0.6B & 29.27 & 16.65 & 11.41 & 8.66 & 23.62  \\
    \bottomrule
    \end{tabular}
}
\caption{Impact of LLM selection.}
\label{tab:llm_selection}
\end{table}

\subsection{Temporal Sampling Strategies}
\label{subsec:temporal}
We further investigate the impact of various temporal sampling strategies, as summarized in Table~\ref{tab:temporal_ablation}. Our results indicate that a stride of $s_e=8$ paired with Single Frame sampling yields the optimal performance. While smaller strides introduce excessive temporal redundancy and noise, larger strides sacrifice fine-grained facial-expression cues indispensable for accurate translation. Furthermore, we evaluate alternative aggregation methods, specifically Max Pooling and Mean Pooling. Single Frame sampling consistently outperforms these pooling-based approaches. This superiority can be attributed to the transient nature of facial expressions; whereas pooling operations tend to attenuate discriminative signals by over-smoothing temporal transitions, direct sampling efficiently preserves the intensity variations of FE.

\begin{table}[!htbp]
\centering
\renewcommand{\arraystretch}{0.8}
\resizebox{\linewidth}{!}{
    \begin{tabular}{lcccccc}
    \toprule
    \textbf{Strategy} & $s_e$ & B-1 & B-2 & B-3 & B-4 & R-L  \\
    \midrule
    Single Frame & 2 & 51.05 & 38.32 & 30.56 & 25.34 & 47.39 \\
    Single Frame & 4 & 51.76 & 39.08 & 31.03 & 25.50 & 47.78 \\
    \textbf{Single Frame} & \textbf{8} & \textbf{52.59} & \textbf{39.72} & \textbf{31.67} & \textbf{26.38} & \textbf{48.63} \\
    Single Frame & 16 & 51.99 & 39.40 & 31.40 & 26.01 & 48.10 \\
    \midrule
    Max Pooling & 8 & 51.48 & 38.45 & 30.50 & 25.36 & 47.14 \\
    Mean Pooling & 8 & 51.83 & 38.90 & 30.91 & 25.54 & 48.34 \\
    \bottomrule
    \end{tabular}
}
\caption{Temporal modeling analysis with different downsampling strategies and step sizes ($s_e$).}
\label{tab:temporal_ablation}
\end{table}

\subsection{Evaluation Protocol for Interrogative Ambiguity Resolution}
\label{subsec:quantitative}

Interrogative constructions provide a rigorous testbed for evaluating ambiguity resolution in MS, since their syntactic interpretation in SL often hinges on NMS rather than MS alone. 
A notable challenge stems from the structural heterogeneity of the CSL-Daily corpus, where interrogative and declarative clauses frequently co-occur within a single utterance. 
For example, the sentence ``这件红色的衣服怎么样？这是新的。'' ("How about this red garment? It's new.") interleaves a wh-question with a subsequent declarative statement. 
Under such conditions, sentence-level classification proves inadequate for capturing fine-grained syntactic transitions, as it cannot ascertain whether the model correctly identifies interrogative cues amid mixed clause types. 
To mitigate this limitation, we introduce a punctuation-sequence evaluation protocol that isolates punctuation streams from both reference and hypothesis translations by removing all lexical tokens. 
Using the above example, the reference punctuation sequence ``？。'' serves as the target against which we compute instance-level Precision, Recall, and F1 scores following the ROUGE-L formulation detailed in Appendix~\ref{sec:metrics_details}. 
F1 scores are averaged at the instance level, and we report macro-averaged results by taking the arithmetic mean across all test instances, ensuring uniform contribution regardless of utterance length or punctuation density. 
This design isolates the model's capacity to generate syntactically appropriate markers under varying contextual conditions while controlling for lexical variation. 

We further curate a subset of 116 interrogative instances from the CSL-Daily test set by filtering for sentences containing question marks. 
As shown in Table~\ref{tab:quantitative}, our method achieves a macro-F1 gain of +3.69 on this subset, suggesting that facial dynamics provide useful cues for recovering interrogative structure.

\begin{table}[htbp]
\centering
\footnotesize
\renewcommand{\arraystretch}{0.8}
\begin{tabular}{lcccc}
\toprule
\textbf{Model} & \textbf{Precision} & \textbf{Recall} & \textbf{F1} \\
\midrule
SpaMo & 67.39 & 64.94 & 64.94 \\
\textbf{FEA-SLT} & \textbf{69.25} & \textbf{69.61} & \textbf{68.63} \\
\bottomrule
\end{tabular}
\caption{Evaluation results of macro-averaged Precision, Recall, and F1 scores for punctuation sequence on the CSL-Daily test set.}
\label{tab:quantitative}
\end{table}

\subsection{Additional Results}
\label{subsec:additional_results}

We provide additional qualitative examples from the PHOENIX14T and CSL-Daily test sets in Tables~\ref{tab:phoenix_results} and~\ref{tab:csl_results}, comparing FEA-SLT outputs with reference translations and reproduced SpaMo \cite{hwang2025spamo} baselines. Color coding indicates translation accuracy: \colorbox{correct}{green} for correct outputs, \colorbox{diffword}{yellow} for semantically equivalent rephrasings, and \colorbox{wrong}{red} for errors; English glosses appear in parentheses for reference.

\begin{figure}[tbp]
\centering
\includegraphics[width=\columnwidth]{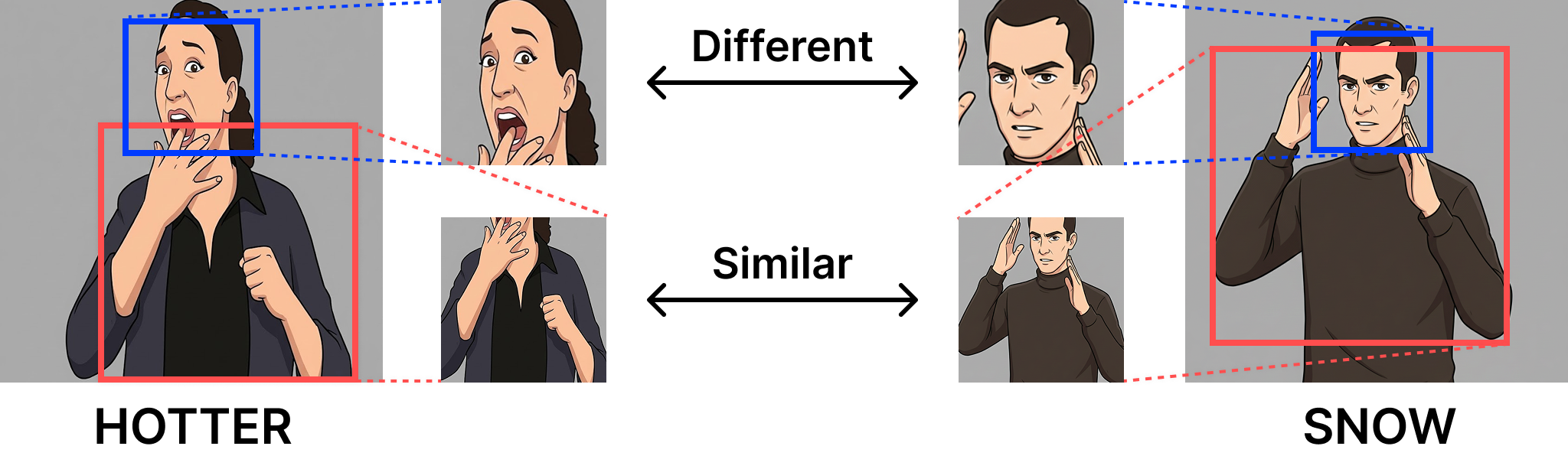}
\caption{Video segments from PHOENIX14T demonstrating almost identical MS for ``HOTTER'' and ``SNOW'', where contrasting facial expressions provide important disambiguating cues.}
\label{fig:snow2hotter}
\end{figure}

On PHOENIX14T (Table~\ref{tab:phoenix_results}), FEA-SLT better preserves temporal and contextual details. In Examples (5), (7), and (9), the baseline occasionally misidentifies temporal expressions, whereas our method aligns more closely with references. Example (2) provides a representative facial-sensitive case: as shown in Figure~\ref{fig:snow2hotter}, the manual components of ``SNOW'' and ``HOTTER'' are visually similar but accompanied by different facial expressions. FEA-SLT matches the reference where the baseline does not, consistent with facial cues providing complementary information in this case.

Results on CSL-Daily (Table~\ref{tab:csl_results}) further show that FEA-SLT better handles interrogative structure in the selected examples. In Examples (3) and (19), the references contain interrogative sentences accompanied by characteristic questioning expressions. SpaMo produces declarative outputs that lose the interrogative intent, whereas FEA-SLT preserves the question structure. Conversely, in Examples (2) and (7), where the references lack questioning markers, FEA-SLT also produces declarative outputs. These cases provide qualitative support for improved facial-sensitive translation; explicit recognition of grammatical facial markers would require dedicated annotations. Additional examples show closer alignment between FEA-SLT outputs and reference meanings, while baseline translations occasionally diverge in logical coherence or semantic intent.

These qualitative observations complement our quantitative results, providing supportive evidence that facial-expression modeling can improve both linguistic content and grammatical structure in selected facial-sensitive cases.

\begin{table*}[t]
\centering
\footnotesize
\setlength{\tabcolsep}{3pt}
\renewcommand{\arraystretch}{1.05}
\setcounter{ablationrow}{0}
\begin{tabular}{@{}>{\centering\arraybackslash}p{0.03\textwidth}>{\raggedright\arraybackslash}p{0.31\textwidth}>{\raggedright\arraybackslash}p{0.31\textwidth}>{\raggedright\arraybackslash}p{0.31\textwidth}@{}}
\toprule
 & \textbf{Reference} & \textbf{SpaMo (Baseline)} & \textbf{FEA-SLT (Ours)} \\
\midrule
\multirow{1}{*}{\centering \ablnum} & in der mitte lockert es auch mal auf & 
in der mitte \colorbox{wrong}{dagegen} \colorbox{wrong}{gebietsweise klar} & 
in der mitte \colorbox{correct}{lockert es auch mal auf} \\
& (In the middle, it also clears up at times.) & 
(\colorbox{wrong}{In contrast,} \colorbox{wrong}{partly clear} in the middle.) & 
(In the middle, \colorbox{correct}{it also clears up} \colorbox{correct}{at times}.) \\
\midrule

\multirow{1}{*}{\centering \ablnum} & samstag fällt auch noch schnee & 
\colorbox{wrong}{und am} samstag \colorbox{wrong}{könnte es dann schon} \colorbox{wrong}{wieder heißer} & 
\colorbox{wrong}{(w/o ``fällt'')} \colorbox{correct}{auch} \colorbox{diffword}{am} samstag \colorbox{correct}{noch schnee} \\
& (On Saturday, it also still snows.) & 
(\colorbox{wrong}{And} on Saturday, \colorbox{wrong}{it could already be} \colorbox{wrong}{hotter again}.) & 
(\colorbox{wrong}{(w/o ``fall'')} \colorbox{correct}{Also snow} on Saturday.) \\
\midrule

\multirow{1}{*}{\centering \ablnum} & in der nacht muss vor allem in der nordwesthälfte mit schauern und gewittern gerechnet werden die heftig ausfallen können & 
\colorbox{wrong}{bevor sich} in der nacht vor allem in der nordwesthälfte \colorbox{diffword}{schauer und gewitter} \colorbox{wrong}{entwickeln können} die \colorbox{wrong}{örtlich auch} \colorbox{diffword}{kräftig sein} können & 
in der nacht muss vor allem in der nordwesthälfte \colorbox{correct}{mit schauern und gewittern} \colorbox{correct}{gerechnet werden} die \colorbox{diffword}{teilweise kräftig} \colorbox{correct}{ausfallen} können \\
& (At night, especially in the northwestern half, showers and thunderstorms must be expected, which can be heavy.) & 
(\colorbox{wrong}{Before} showers and thunderstorms \colorbox{wrong}{can develop} at night, especially in the northwestern half, they can \colorbox{wrong}{also} be \colorbox{wrong}{locally} \colorbox{diffword}{strong}.) & 
(At night, especially in the northwestern half, showers and thunderstorms must \colorbox{correct}{be expected}, which can be \colorbox{diffword}{partly} \colorbox{diffword}{strong}.) \\
\midrule

\multirow{1}{*}{\centering \ablnum} & im westen und nordwesten fallen einzelne schauer & 
im westen und nordwesten \colorbox{diffword}{gibt es} einzelne schauer & 
im westen und nordwesten \colorbox{correct}{fallen} einzelne schauer \\
& (In the west and northwest, isolated showers fall.) & 
(In the west and northwest, \colorbox{diffword}{there are} isolated showers.) & 
(In the west and northwest, isolated showers \colorbox{correct}{fall}.) \\
\midrule
\multirow{1}{*}{\centering \ablnum} & und nun die wettervorhersage für morgen samstag den sechsundzwanzigsten januar & 
und nun die wettervorhersage für morgen samstag den sechsundzwanzigsten \colorbox{wrong}{juni} & 
und nun die wettervorhersage für morgen samstag den sechsundzwanzigsten \colorbox{correct}{januar} \\
& (And now the weather forecast for tomorrow, Saturday the 26th of January.) & 
(And now the weather forecast for tomorrow, Saturday the 26th of \colorbox{wrong}{June}.) & 
(And now the weather forecast for tomorrow, Saturday the 26th of \colorbox{correct}{January}.) \\
\midrule

\multirow{1}{*}{\centering \ablnum} & auch in den folgenden tagen ändert sich an diesem wechselhaften wetter wenig & 
\colorbox{wrong}{(w/o ``auch'')} in den folgenden tagen \colorbox{wrong}{bleibt es immer noch} \colorbox{diffword}{wechselhaft} \colorbox{wrong}{und} \colorbox{wrong}{nicht mehr ganz so windig} & 
\colorbox{correct}{auch} in den folgenden tagen \colorbox{correct}{ändert sich} \colorbox{correct}{an} \colorbox{diffword}{dem} \colorbox{correct}{wechselhaften wetter wenig} \\
& (Also in the following days, this changeable weather changes little.) & 
(\colorbox{wrong}{(w/o ``Also'')} In the following days, \colorbox{wrong}{it remains} changeable \colorbox{wrong}{and no longer quite so windy}.) & 
(\colorbox{correct}{Also} in the following days, \colorbox{correct}{this} changeable \colorbox{correct}{weather changes little}.) \\
\midrule

\multirow{1}{*}{\centering \ablnum} & und nun die wettervorhersage für morgen mittwoch den neunzehnten januar & 
und nun die wettervorhersage für morgen mittwoch den neunzehnten \colorbox{wrong}{dezember} & 
und nun die wettervorhersage für morgen mittwoch den neunzehnten \colorbox{correct}{januar} \\
& (And now the weather forecast for tomorrow, Wednesday the 19th of January.) & 
(And now the weather forecast for tomorrow, Wednesday the 19th of \colorbox{wrong}{December}.) & 
(And now the weather forecast for tomorrow, Wednesday the 19th of \colorbox{correct}{January}.) \\
\midrule

\multirow{1}{*}{\centering \ablnum} & ich wünsche ihnen noch einen schönen abend & 
\colorbox{wrong}{(w/o ``ich wünsche ihnen noch einen'')} schönen abend noch & 
\colorbox{wrong}{und jetzt} \colorbox{correct}{wünsche ich ihnen noch} einen schönen abend \\
& (I wish you a nice evening.) & 
(\colorbox{wrong}{(w/o ``I wish you'')} Have a nice evening.) & 
(\colorbox{wrong}{And now} \colorbox{correct}{I wish you} a nice evening.) \\
\midrule

\multirow{1}{*}{\centering \ablnum} & und nun die wettervorhersage für morgen freitag den neunten oktober & 
und nun die wettervorhersage für morgen freitag den \colorbox{wrong}{achten} oktober & 
und nun die wettervorhersage für morgen freitag den \colorbox{correct}{neunten} oktober \\
& (And now the weather forecast for tomorrow, Friday the 9th of October.) & 
(And now the weather forecast for tomorrow, Friday the \colorbox{wrong}{8th} of October.) & 
(And now the weather forecast for tomorrow, Friday the \colorbox{correct}{9th} of October.) \\
\midrule

\multirow{1}{*}{\centering \ablnum} & auf den bergen sind orkanartige böen möglich & 
auf den bergen \colorbox{diffword}{kann es} \colorbox{wrong}{bodenfrost} \colorbox{diffword}{geben} & 
auf den bergen \colorbox{diffword}{kann es} \colorbox{correct}{orkanartige böen} \colorbox{diffword}{geben} \\
& (On the mountains, hurricane-like gusts are possible.) & 
(On the mountains, \colorbox{diffword}{there can be} \colorbox{wrong}{ground frost}.) & 
(On the mountains, \colorbox{diffword}{there can be} \colorbox{correct}{hurricane-like gusts}.) \\
\midrule

\multirow{1}{*}{\centering \ablnum} & abseits der gewitter weht der wind schwach bis mäßig, an der küste frisch & 
\colorbox{wrong}{bei} gewittern weht der wind schwach bis mäßig, an \colorbox{diffword}{den küsten} \colorbox{wrong}{mäßig} & 
\colorbox{correct}{abseits der} gewitter weht der wind schwach bis mäßig, an \colorbox{diffword}{den küsten} \colorbox{wrong}{auch} \colorbox{correct}{frisch} \\
& (Away from thunderstorms, the wind blows light to moderate, fresh at the coast.) & 
(\colorbox{wrong}{During} thunderstorms, the wind blows light to moderate, at \colorbox{diffword}{the coasts} \colorbox{wrong}{moderate}.) & 
(\colorbox{correct}{Away from} thunderstorms, the wind blows light to moderate, at \colorbox{diffword}{the coasts} \colorbox{wrong}{also} \colorbox{correct}{fresh}.) \\
\midrule

\multirow{1}{*}{\centering \ablnum} & am tag vor allem im norden regen & 
\colorbox{wrong}{(w/o ``am tag'')} vor allem im norden \colorbox{diffword}{regnet es} & 
\colorbox{correct}{am tag} vor allem im norden \colorbox{correct}{regen} \\
& (It rains especially in the north during the day.) & 
(It \colorbox{diffword}{is raining} especially in the north \colorbox{wrong}{(w/o ``during the day'')}.) & 
(It \colorbox{correct}{rains} especially in the north \colorbox{correct}{during the day}.) \\

\bottomrule
\end{tabular}
\caption{More qualitative results on PHOENIX14T showing FEA-SLT's advantages in precise details.}
\label{tab:phoenix_results}
\end{table*}

\begin{table*}[t]
\centering
\footnotesize
\setlength{\tabcolsep}{3pt}
\renewcommand{\arraystretch}{1.05} 
\setcounter{ablationrow}{0}
\begin{tabular}{@{}>{\centering\arraybackslash}p{0.03\textwidth}>{\raggedright\arraybackslash}p{0.31\textwidth}>{\raggedright\arraybackslash}p{0.31\textwidth}>{\raggedright\arraybackslash}p{0.31\textwidth}@{}}
\toprule
 & \textbf{Reference} & \textbf{SpaMo (Baseline)} & \textbf{FEA-SLT (Ours)} \\
\midrule

\multirow{1}{*}{\centering \ablnum} & 我给你预订了一个生日蛋糕。 & 
我\colorbox{wrong}{去火车站送}生日蛋糕。 & 
我\colorbox{correct}{预订}一个生日蛋糕。 \\
& (I booked a birthday cake for you.) & 
(I \colorbox{wrong}{went to the train station to deliver} a birthday cake.) & 
(I \colorbox{correct}{booked} a birthday cake \colorbox{wrong}{(w/o ``for you'')}.) \\
\midrule

\multirow{1}{*}{\centering \ablnum} & 今天是阴天,可能会下雨。 & 
今天\colorbox{wrong}{下雨},会下雨\colorbox{wrong}{吗?} & 
今天\colorbox{correct}{阴天},\colorbox{diffword}{可能}下雨\colorbox{correct}{。} \\
& (Today is overcast; it might rain.) & 
(It's \colorbox{wrong}{raining} today; \colorbox{wrong}{will} it rain \colorbox{wrong}{?}) & 
(Today is \colorbox{correct}{overcast}; rain is \colorbox{diffword}{possible}\colorbox{correct}{.}) \\
\midrule

\multirow{1}{*}{\centering \ablnum} & 微信是谁发的。 & 
\colorbox{wrong}{是,给你}发\colorbox{diffword}{个}微信。 & 
微信是\colorbox{correct}{谁}发\colorbox{correct}{的}\colorbox{diffword}{?} \\
& (Who sent the WeChat message?) & 
(\colorbox{wrong}{Yes,} send \colorbox{wrong}{you a} WeChat message\colorbox{wrong}{.}) & 
(\colorbox{correct}{Who} sent the WeChat message\colorbox{correct}{?}) \\
\midrule

\multirow{1}{*}{\centering \ablnum} & 我的笔记本电脑不见了! & 
我\colorbox{wrong}{把}笔记本电脑\colorbox{wrong}{拆了。} & 
我的笔记本电脑\colorbox{correct}{不见了}\colorbox{wrong}{。} \\
& (My laptop is missing!) & 
(\colorbox{wrong}{I disassembled} my laptop\colorbox{wrong}{.}) & 
(My laptop \colorbox{correct}{is missing}\colorbox{wrong}{.}) \\
\midrule

\multirow{1}{*}{\centering \ablnum} & 他参加了图书馆的暑期活动。 & 
他\colorbox{wrong}{在}图书馆\colorbox{wrong}{定期锻炼}。 & 
他\colorbox{correct}{参加了}图书馆\colorbox{correct}{的暑期活动}。 \\
& (He participated in the library's summer activities.) & 
(He \colorbox{wrong}{exercises regularly at} the library.) & 
(He \colorbox{correct}{participated in} the library\colorbox{correct}{'s} \colorbox{correct}{summer activities}.) \\
\midrule

\multirow{1}{*}{\centering \ablnum} & 我们做人不能自私。 & 
我们做人\colorbox{diffword}{,}不能\colorbox{wrong}{欺负人}。 & 
我们做人不能\colorbox{correct}{自私}\colorbox{diffword}{自利}。 \\
& (We must not be selfish as human beings.) & 
(We must not \colorbox{wrong}{bully others} as human beings.) & 
(We must not \colorbox{correct}{be selfish} \colorbox{diffword}{and self-serving} as human beings.) \\
\midrule


\multirow{1}{*}{\centering \ablnum} & 这个月我每天游泳。 & 
\colorbox{wrong}{为什么}我每天\colorbox{diffword}{去}游泳\colorbox{wrong}{?} & 
\colorbox{wrong}{下}\colorbox{correct}{个月}我每天游泳\colorbox{correct}{。} \\
& (This month, I swim every day.) & 
(\colorbox{wrong}{Why do} I \colorbox{diffword}{go} swimming every day\colorbox{wrong}{?}) & 
(\colorbox{wrong}{Next} month, I swim every day\colorbox{correct}{.}) \\
\midrule


\multirow{1}{*}{\centering \ablnum} & 目前许多笔记本电脑都内置了蓝牙功能。 & 
笔记本电脑\colorbox{wrong}{和蓝牙电脑一起使用}。 & 
\colorbox{diffword}{很多}笔记本电脑\colorbox{correct}{都}\colorbox{diffword}{配有}\colorbox{correct}{蓝牙功能}。 \\
& (Currently, many laptops have built-in Bluetooth functionality.) & 
(Laptops \colorbox{wrong}{and Bluetooth computers} \colorbox{wrong}{are used together}.) & 
(\colorbox{correct}{Many} laptops \colorbox{diffword}{are equipped with} \colorbox{correct}{Bluetooth functionality}.) \\
\midrule

\multirow{1}{*}{\centering \ablnum} & 你和小张什么时候认识的? & 
\colorbox{wrong}{你们}什么时候认识? & 
\colorbox{correct}{你和小张}什么时候认识? \\
& (When did you and Xiao Zhang meet?) & 
(When did \colorbox{wrong}{you} meet?) & 
(When did \colorbox{correct}{you and Xiao Zhang} meet?) \\
\midrule

\multirow{1}{*}{\centering \ablnum} & 他们想什么时候去买椅子? & 
\colorbox{wrong}{你们}什么时候去买椅子? & 
\colorbox{correct}{他们想}什么时候去买椅子? \\
& (When do they want to go buy chairs?) & 
(When do \colorbox{wrong}{you} go buy chairs?) & 
(When do \colorbox{correct}{they want to} go buy chairs?) \\
\midrule

\multirow{1}{*}{\centering \ablnum} & 时间过得真快,新年快要到了。 & 
时间\colorbox{wrong}{越快越好}。 & 
时间\colorbox{correct}{过得快},\colorbox{correct}{新年快到了}。 \\
& (Time flies so fast; the New Year is almost here.) & 
(\colorbox{wrong}{The faster} time \colorbox{wrong}{passes, the better}.) & 
(Time \colorbox{correct}{flies fast}; \colorbox{correct}{the New Year is} \colorbox{diffword}{coming}\colorbox{correct}{.}) \\
\midrule

\multirow{1}{*}{\centering \ablnum} & 封面设计是一种艺术设计。 & 
\colorbox{wrong}{艺术}设计是一种\colorbox{wrong}{技术}设计。 & 
\colorbox{correct}{封面}设计是一种\colorbox{correct}{艺术}设计。 \\
& (Cover design is a type of artistic design.) & 
(\colorbox{wrong}{Artistic} design is a type of \colorbox{wrong}{technical} design.) & 
(\colorbox{correct}{Cover} design is a type of \colorbox{correct}{artistic} design.) \\
\midrule

\multirow{1}{*}{\centering \ablnum} & 会议室里有多媒体系统。 & 
\colorbox{wrong}{国会}有\colorbox{diffword}{一个}\colorbox{wrong}{拨款}系统。 & 
\colorbox{correct}{会议室}有\colorbox{correct}{多媒体}系统。 \\
& (There is a multimedia system in the conference room.) & 
(\colorbox{wrong}{Congress} has \colorbox{diffword}{an} \colorbox{wrong}{appropriations} system.) & 
(There is a \colorbox{correct}{multimedia} system \colorbox{correct}{in the conference room}.) \\
\midrule

\multirow{1}{*}{\centering \ablnum} & 我们不想毕业。 & 
我们不想\colorbox{wrong}{上大学}。 & 
我们不想\colorbox{correct}{毕业}。 \\
& (We don't want to graduate.) & 
(We don't want to \colorbox{wrong}{go to college}.) & 
(We don't want to \colorbox{correct}{graduate}.) \\
\midrule

\multirow{1}{*}{\centering \ablnum} & 这场比赛不公平! & 
这场比赛\colorbox{wrong}{很公平}\colorbox{wrong}{。} & 
这场比赛\colorbox{correct}{不公平}\colorbox{wrong}{。} \\
& (This match is unfair!) & 
(This match is \colorbox{wrong}{very fair}\colorbox{wrong}{.}) & 
(This match is \colorbox{correct}{unfair}\colorbox{wrong}{.}) \\
\midrule

\multirow{1}{*}{\centering \ablnum} & 我明天去超市买面包给爷爷吃。 & 
我明天去超市买面包\colorbox{wrong}{,不喜欢}。 & 
我明天去超市买\colorbox{diffword}{点}面包\colorbox{correct}{给爷爷}。 \\
& (I will go to the supermarket tomorrow to buy bread for grandpa to eat.) & 
(I will go to the supermarket tomorrow to buy bread\colorbox{wrong}{, don't like it}.) & 
(I will go to the supermarket tomorrow to buy \colorbox{diffword}{some} bread \colorbox{correct}{for grandpa} \colorbox{wrong}{(w/o "to eat")}.) \\
\midrule

\multirow{1}{*}{\centering \ablnum} & 我的丈夫在美国,他喜欢中国菜。 & 
我\colorbox{wrong}{妻子}\colorbox{diffword}{住}在美国,\colorbox{wrong}{她}喜欢中国菜。 & 
我的\colorbox{correct}{丈夫}在美国,\colorbox{wrong}{(w/o ``他'')}喜欢吃中国菜。 \\
& (My husband is in the US; he likes Chinese food.) & 
(My \colorbox{wrong}{wife} \colorbox{diffword}{lives} in the US; \colorbox{wrong}{she} likes Chinese food.) & 
(My \colorbox{correct}{husband} \colorbox{correct}{is} in the US; \colorbox{wrong}{(w/o ``he'')}likes Chinese food.) \\
\midrule

\multirow{1}{*}{\centering \ablnum} & 钥匙坏了,拿不出来了。 & 
\colorbox{wrong}{这太暴力了,我不能忍受}。 & 
\colorbox{diffword}{这把}\colorbox{correct}{钥匙坏了},\colorbox{diffword}{我}\colorbox{correct}{拿不出来}\colorbox{correct}{。} \\
& (The key is broken; I can't take it out.) & 
(\colorbox{wrong}{This is too violent; I can't stand it}.) & 
(\colorbox{diffword}{This} \colorbox{correct}{key is broken}; \colorbox{diffword}{I} \colorbox{correct}{can't take it out}.) \\
\midrule

\multirow{1}{*}{\centering \ablnum} & 椅子上有一件衣服,是谁的? & 
\colorbox{wrong}{桌子}{上有一}\colorbox{diffword}{只}\colorbox{wrong}{手表。} & 
\colorbox{wrong}{这个桌子}上有一\colorbox{correct}{件衣服},\colorbox{correct}{是谁的?} \\
& (There is a piece of clothing on the chair; whose is it?) & 
(There is a \colorbox{wrong}{watch} on \colorbox{correct}{the} \colorbox{wrong}{table.}) & 
(There is a \colorbox{correct}{piece of clothing} on \colorbox{wrong}{this table}; \colorbox{correct}{whose is it?}) \\

\bottomrule
\end{tabular}
\caption{More qualitative results on CSL-Daily showing FEA-SLT's advantages in interrogative structures, associated facial expressions, and translation quality.}
\label{tab:csl_results}
\end{table*}

\end{CJK}
\end{document}